\documentclass{article} 
\usepackage{iclr2027_conference,times}

\usepackage{amsmath,amsfonts,bm}

\def\eqref#1{equation~\ref{#1}}

\def\1{\bm{1}}

\DeclareMathAlphabet{\mathsfit}{\encodingdefault}{\sfdefault}{m}{sl}
\SetMathAlphabet{\mathsfit}{bold}{\encodingdefault}{\sfdefault}{bx}{n}

\usepackage{hyperref}
\usepackage{url}

\usepackage{amsmath,amsfonts}
\usepackage{amssymb}
\usepackage{array}
\usepackage{graphicx}
\usepackage{multirow}
\usepackage{booktabs}
\usepackage[table]{xcolor}
\usepackage{arydshln}
\usepackage{colortbl}
\usepackage{orcidlink}
\usepackage[caption=false,font=normalsize,labelfont=sf,textfont=sf]{subfig}
\usepackage{algorithm}
\usepackage{algorithmic}
\usepackage{capt-of}
\usepackage{tcolorbox}
\tcbuselibrary{breakable}

\newtcolorbox{promptbox}[1]{
    breakable,
    colback=black!2,
    colframe=black!25,
    colbacktitle=black!6,
    coltitle=black,
    fonttitle=\bfseries\small,
    fontupper=\small,
    boxrule=0.4pt,
    arc=0pt,
    left=8pt,right=8pt,top=6pt,bottom=6pt,
    before skip=8pt,after skip=8pt,
    title={#1},
    title after break={#1 (continued)}
}
\definecolor{lighty}{RGB}{252,245,222}
\definecolor{lightg}{RGB}{232,245,232}

\title{Writer-R1: Dynamic Rubric Mining for Reinforcement Learning of Open-Ended Writing}

\author{%
\textbf{Jihao Zhao\textsuperscript{1},
Shuaishuai Zu\textsuperscript{1},
Zhiyuan Ji\textsuperscript{1},
Chunlai Zhou\textsuperscript{1},
Biao Qin\thanks{Corresponding author}\hspace{0.6em}\textsuperscript{1}
}\\
\textsuperscript{1} School of Information, Renmin University of China, Beijing, China
}

\iclrfinalcopy
\begin{document}

\maketitle
\lhead{Preprint}

\begin{abstract}
Training models for open-ended writing requires diverse prompts and sufficiently precise specifications of response quality. Constructing such specifications at scale entails recovering task requirements, consolidating overlapping criteria, and defining observable differences between performance levels. We introduce Writer-R1, a writing training pipeline built around Inductive Rubric Generation (IRG), which integrates these operations into an automated rubric construction procedure. The pipeline expands an open writing corpus and uses the resulting prompt-specific rubrics throughout supervised training and reinforcement learning (RL). Each rubric is concatenated with the user request and also defines the reward criteria, making the requirements used for assessment available to the writer before generation. A compact model, RubricGen-4B, learns to supply rubrics for new requests. Writer-R1 improves WritingBench scores by 0.182 and 0.158 over the 4B and 8B backbones, respectively, with gains also on HelloBench. Training ablations favor IRG over direct rubric generation, and further analyses support rubric transfer and the robustness of writer improvements across evaluators. These results support using inductively mined rubrics as both generation guidance and reward specifications for open-ended writing. Our code is available at \url{https://github.com/Robot2050/Writer-R1}.
\end{abstract}

\section{Introduction}
Open-ended writing requires large language models (LLMs) to coordinate content, discourse organization, and task-specific constraints~\citep{deng2022model,pham2024suri}. Synthetic long-output supervision extends generation length~\citep{bai2024longwriter}, while RL further improves writing quality and length control~\citep{wu2025longwriter}. WritingBench and HelloBench assess performance across domains, formats, and length requirements, demonstrating the need to evaluate several aspects of writing quality~\citep{wu2025writingbench,que2024hellobench}. Training a capable writer therefore requires specifying what makes a response successful for its particular request, while accommodating multiple valid arguments, structures, and styles.

Reward design must translate these varying requirements into supervision. Human feedback improves generation beyond demonstration learning~\citep{stiennon2020learning,ouyang2022training}, but dependable judgments require careful annotation design~\citep{van2019best}. Automatic evaluation captures distributional properties of generated text~\citep{pillutla2021mauve}, while studies of conventional metrics identify limitations in their agreement with human judgments~\citep{novikova2017we,chhun2022human}. LLM evaluators support explicit assessment criteria~\citep{liu2023g}, although their reliability depends on evaluation instructions and judgment biases~\citep{zheng2023judging,gu2024survey}. Rubrics formalize the assessment basis by specifying dimensions and levels of satisfaction, as in the query-dependent criteria used by WritingBench~\citep{wu2025writingbench}. Recent work applies rubrics to RL for open-ended responses~\citep{gunjal2026rubrics,wei2026qurl}, synthesizes rubric-annotated data at scale~\citep{li2026ares}, and constructs reusable prompt-level reward specifications~\citep{weng2026specifications}.

For writing, these developments raise a coupled construction and training problem. A useful rubric must recover explicit constraints and implicit task requirements, avoid counting the same property repeatedly, and distinguish substantive improvements from superficial compliance. Broad criteria such as clarity or argument quality remain underspecified until they identify the evidence that should be present in the requested text. Repeated optimization makes these choices consequential: omitted requirements receive no direct reward, redundant criteria alter their effective weighting, and vague descriptions leave scoring dependent on evaluator interpretation. Scaling training also requires a sufficiently varied prompt collection. WritingBench releases its benchmark and evaluation tools, but not the writing corpus used to train its reported writing model. A practical workflow must therefore construct diverse requests together with the specifications needed to select demonstrations and evaluate new responses.

We introduce \textbf{Writer-R1}, a writing training pipeline centered on \textbf{IRG}. IRG combines requirement discovery, dimension consolidation, and scoring specification to construct explicit criteria from each writing request. Discovery prioritizes coverage; consolidation organizes related requirements into complementary assessment targets; scoring descriptions make differences in satisfaction observable. We apply this procedure to reconstructed and enriched LongWrite prompts~\citep{wu2025longwriter}, obtaining a corpus that supports both supervised fine-tuning (SFT) and RL. Each rubric is constructed before policy optimization and reused across sampled responses. Dynamic thus refers to adaptation across writing requests, with a stable specification for each training prompt.

A central design choice is to concatenate the rubric with the user request. The writer can then condition on the same requirements that govern demonstration selection and reward computation. This explicit conditioning directs generation toward task-relevant properties while preserving flexibility in how they are expressed. RubricGen-4B learns the mapping from requests to mined rubrics and supplies this guidance for new inputs at inference. The resulting workflow makes a common task specification available to the writer and its evaluator throughout learning and generation. With Qwen3-4B, IRG exceeds direct rubric generation by 0.097 on WritingBench after both SFT and SFT+RL. Each construction method supplies its own supervision and inference rubrics under matched downstream procedures. Writer-R1 also improves over both backbones on HelloBench. Further analyses test whether mined rubrics guide other writers and whether the improvements persist across automated and human assessment.

Our contributions are threefold:
\begin{itemize}
    \item We introduce IRG, an integrated procedure that discovers writing requirements, consolidates assessment dimensions, and specifies observable performance levels. The mined rubrics jointly provide generation guidance and RL reward criteria.
    \item We develop Writer-R1 to learn from these rubrics through SFT and RL, construct 19K training instances from LongWrite, and train RubricGen-4B to supply task-specific rubrics for new requests.
    \item We demonstrate writing gains on WritingBench and HelloBench and an advantage for IRG over direct generation in matched training comparisons. Studies with six target writers, four LLM evaluators, and four experts further assess rubric transfer and writer robustness.
\end{itemize}

\section{Related Work}
\label{sec:related}
\textbf{Training models for writing.}
LongWriter~\citep{bai2024longwriter} extends generation length through synthetic long-output supervision, and subsequent work studies self-generated length supervision~\citep{quan2024language}. Suri~\citep{pham2024suri} combines complex instructions with instruction-based preference optimization to improve constraint satisfaction. Planning-based approaches address discourse organization through plot and outline induction~\citep{yao2019plan,lee2025navigating}, recursive planning~\citep{xiong2025beyond}, and reflection during generation~\citep{wu2025superwriter}. WritingBench~\citep{wu2025writingbench} and LongGenBench~\citep{wu2024longgenbench} broaden evaluation across writing requirements and long-form generation. LongWriter-Zero~\citep{wu2025longwriter} trains with rewards for quality, length, and formatting. Writer-R1 complements these approaches by constructing a corpus of task-specific rubrics and making them available to the writer during both training and inference.

\textbf{Rewards and rubric-based training.}
Human comparisons support reward learning for generation~\citep{ziegler2019fine,stiennon2020learning} and instruction following~\citep{ouyang2022training}, while direct preference optimization provides an alternative policy objective~\citep{rafailov2023direct}. Mathematical answer verification~\citep{shao2024deepseekmath,guo2025deepseek} and code execution~\citep{le2022coderl,liu2023rltf} provide more direct reward signals. Further work uses compiler feedback~\citep{dou2024stepcoder} and supervision on intermediate reasoning steps~\citep{zhang2025prm}. Open-ended writing requires judgments about several interacting qualities, motivating LLM assessment~\citep{lee2023rlaif,liu2023g} and explicit rubrics. WritingBench generates query-dependent criteria and scoring descriptions for evaluation and data selection. Rubrics as Rewards~\citep{gunjal2026rubrics} studies how rubric feedback supports RL in medical and scientific tasks. QuRL~\citep{wei2026qurl} mines question-specific rubrics from online sources, ARES~\citep{li2026ares} synthesizes rubric-annotated training instances, and Prompt-Level Reward Specifications~\citep{weng2026specifications} constructs reusable criteria before policy training. IRG draws on inductive category development~\citep{glaser2017discovery,corbin2014basics} to organize writing requirements, then defines their assessment scopes and performance levels. Writer-R1 uses the resulting rubrics as both generation guidance and reward specifications. The training ablations evaluate this integrated construction procedure.

\section{Dynamic Rubric Mining for Writer-R1}
\label{sec:method}
\subsection{Overview and Rubric Representation}
\begin{figure}[t]
    \centering
    \includegraphics[width=\textwidth]{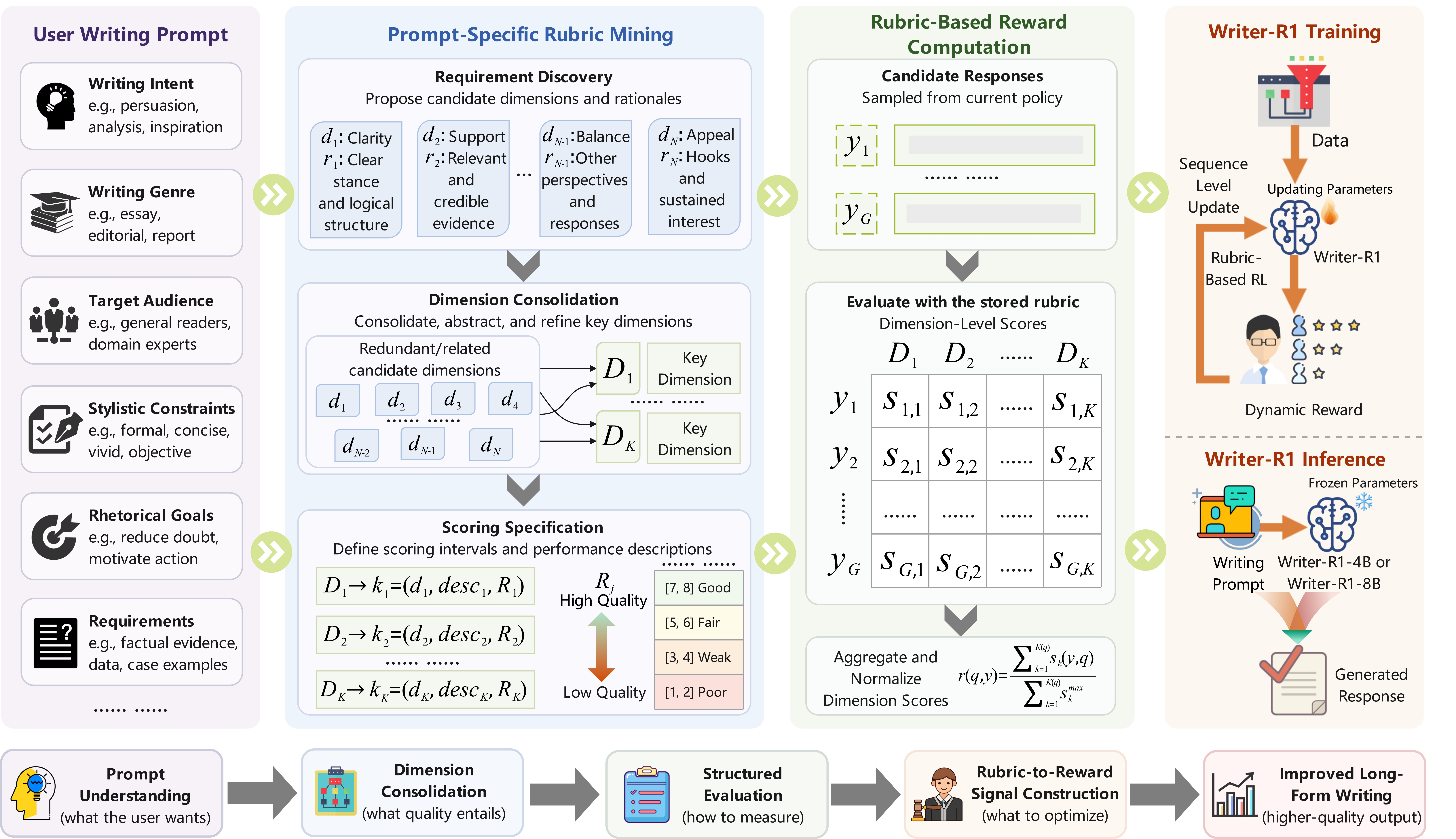}
    \caption{Writer-R1 workflow. IRG rubrics condition generation and define RL rewards; RubricGen-4B supplies inference-time rubrics. $N$, $K$, and $G$ count candidate dimensions, retained dimensions, and sampled responses; $S_{i,k}$ scores response $i$ on dimension $k$.}
    \label{fig:kuangjia}
\end{figure}

Writer-R1 constructs writing prompts with task-specific rubrics, then uses each rubric as both generation context and a specification for scoring responses (Figure~\ref{fig:kuangjia}). For a prompt $q$, the rubric is
\begin{equation}
\label{eq:rubric}
\mathcal{R}(q)=\{(d_k,a_k,\mathcal{B}_k)\}_{k=1}^{K(q)},
\end{equation}
where $d_k$ names a dimension, $a_k$ specifies its assessment scope, and $\mathcal{B}_k$ contains score intervals with performance descriptions. The dimensions, their number $K(q)$, and their score descriptions depend on the prompt. The writer receives $x_q=q\oplus\operatorname{serialize}(\mathcal{R}(q))$, where $\oplus$ denotes direct text concatenation of the user request and its rubric. This makes the intended requirements available before generation, while the rubric describes acceptable properties rather than prescribing a reference response.

The rubric varies across prompts and remains fixed across responses to the same training prompt. Reusing the prompt and rubric pair preserves the assessment basis across policy updates and amortizes construction over repeated scoring. Scope descriptions distinguish required properties from optional stylistic choices, allowing different arguments, structures, and wording to satisfy the same specification. The rubric links generation and evaluation by exposing task requirements to the writer and using those requirements to score its responses.

\subsection{Inductive Rubric Generation}
\label{sec:mining}
IRG constructs the assessment structure implied by a writing request: the relevant quality dimensions, their assessment scopes, and the evidence distinguishing levels of satisfaction. The dimensions are induced from communicative intent, genre conventions, audience expectations, and explicit constraints in the prompt. Three linked LLM operations organize these requirements into a rubric that supports both generation guidance and reward computation.

\textbf{Requirement discovery.}
The constructor is instructed to propose 15 to 20 candidate dimensions, each accompanied by a rationale identifying its assessment target, relevance to the request, and observable evidence of satisfaction. At least half should address query-specific requirements; broader dimensions must also be tied to the task objectives. Discovery prioritizes coverage while discouraging synonymous repetition. The candidates are subsequently consolidated under the same request:
\begin{equation}
\label{eq:mining}
\mathcal{D}^{(0)}(q)=A_{\mathrm{disc}}(q),\qquad
\mathcal{D}^{(1)}(q)=A_{\mathrm{cons}}\big(q,\mathcal{D}^{(0)}(q)\big).
\end{equation}

\textbf{Dimension consolidation.}
Given $q$ and $\mathcal{D}^{(0)}(q)$, the constructor targets 5 to 8 core dimensions through selection and semantic consolidation. Selection prioritizes relevance to task success, observability in the response, and information contributed beyond other dimensions. Candidates with overlapping assessment targets are merged under a unified scope, while requirements with distinct evaluative value are retained. Each output pairs a dimension name $d_k$ with a scope description $a_k$ that preserves the essential requirements of its constituent candidates. This balances coverage against redundancy: omitted requirements lack an explicit assessment criterion, whereas duplicated targets can receive disproportionate weight in the aggregate reward.

\textbf{Scoring specification.}
The final operation constructs $\mathcal{R}(q)=A_{\mathrm{spec}}(q,\mathcal{D}^{(1)}(q))$. A separate call conditioned on $q$, $d_k$, and $a_k$ requests 5 to 8 consecutive, non-overlapping score intervals, each containing two consecutive integer scores. Each interval receives a performance description within the fixed assessment scope. The descriptions must progress in satisfaction level, distinguish adjacent levels through observable evidence, and span clear failure through full satisfaction. For argument quality, these distinctions may concern whether evidence is relevant, whether it supports the conclusion, and whether the inferential connection is explained. The dimension scope remains fixed while the degree of satisfaction changes.

Separating these operations allows the candidate set to emphasize requirement coverage before consolidation determines the granularity of assessment. The final descriptions then specify which differences between responses warrant different scores. This ordering connects prompt interpretation to the structure of the learning signal: dimension selection determines what is assessed, and performance descriptions determine how satisfaction is judged. Concatenating the resulting rubric with the request exposes these same distinctions to the writer before generation. The integrated procedure is evaluated through the training comparisons. Appendix~\ref{app:mining} provides implementation details; Appendix~\ref{app:prompts} collects the operational prompts.

\subsection{Constructing the Prompt and Rubric Corpus}
\label{sec:data}
The 6K prompts in the LongWrite training set~\citep{wu2025longwriter} provide a limited pool for heterogeneous writing requirements. We expand them with Qwen3-32B~\citep{yang2025qwen3}, allocating reconstruction calls equally to two strategies. Both replace wording, scenarios, and source material while preserving task type, language, and comparable difficulty; the second additionally enriches shortened reconstructions to restore context and detail.

We filter degenerate repetition and manually review task clarity, contextual coherence, constraint consistency, and usable length. Training candidates are screened against WritingBench and HelloBench using contiguous trigrams of Qwen3-4B tokenizer tokens. A candidate is discarded when shared trigrams cover at least 10\% of those in any compared test sample, using that sample as the denominator. Very few candidates meet this threshold. The final corpus contains 19K prompts, with rubrics mined by Qwen3-32B, which supplies all generated training content.

RubricGen-4B is trained on all 19K prompt and rubric pairs to learn a direct mapping from requests to the outputs of IRG. For a new request, it predicts $\widehat{\mathcal{R}}(q)$, which is concatenated with $q$ for Writer-R1 generation. The user supplies only the writing request.

\subsection{Rubric-Conditioned Training}
\label{sec:training}
\textbf{Supervised initialization.}
For each of the 15K SFT prompts, Qwen3-32B generates five candidate responses. We select the highest-scoring candidate under the associated rubric as the target $y^\star$. For a target of $T$ tokens, the writer minimizes
\begin{equation}
\label{eq:sft-loss}
\mathcal{L}_{\mathrm{SFT}}(\theta;q,y^\star)
=-\frac{1}{T}\sum_{t=1}^{T}\log\pi_\theta\big(y_t^\star\mid x_q,y_{<t}^\star\big),
\end{equation}
where $x_q$ concatenates the request and rubric, and $y_{<t}^\star$ is the preceding target prefix. The loss is evaluated on the response tokens, with the rubric supplied as conditioning context. Rubric-based selection pairs explicit requirements with a demonstration chosen for its satisfaction of those requirements. SFT therefore initializes the writer to generate complete texts under the same input structure used in RL.

RL continues from this policy on the remaining approximately 4K prompts, comparing newly sampled responses under each prompt's stored rubric. The transition preserves the rubric representation while changing the learning signal from imitation of a selected response to relative assessment of alternative responses. This allows policy optimization to explore alternative responses under the rubric-conditioned formulation learned in SFT.

\textbf{Reward computation.}
Qwen3-32B evaluates each response $y$ against the stored rubric and assigns a score $s_k(y,q)$ to dimension $k$. We aggregate these scores as
\begin{equation}
\label{eq:reward}
r(q,y)=\frac{\sum_{k=1}^{K(q)}s_k(y,q)}{\sum_{k=1}^{K(q)}s_k^{\max}},
\end{equation}
where $s_k^{\max}$ is the maximum for dimension $k$. This is a weighted average of normalized dimension scores, with weights proportional to their maxima. Responses to the same prompt share both the criteria and the denominator. Dimension selection and score ranges therefore jointly determine the reward structure; normalization fixes the maximum aggregate score without equalizing difficulty across prompts. Aggregation reduces dimension scores to a scalar reward, while rubric conditioning preserves the underlying criteria in the policy context. The writer therefore receives explicit guidance before sampling, and the reward subsequently evaluates the resulting text. This connects the assessment structure recovered by IRG to both response generation and policy optimization.

\textbf{Group-based policy updates.}
For each context $x_q$, the behavior policy $\pi_{\theta_{\mathrm{old}}}$ samples $G$ responses. Their rewards define group-relative advantages:
\begin{equation}
\label{eq:advantage}
\widehat{A}_i=\frac{r(q,y_i)-\operatorname{mean}_{j}r(q,y_j)}
{\operatorname{std}_{j}r(q,y_j)+\delta},
\end{equation}
where statistics are computed over the group and $\delta>0$ stabilizes the denominator. Dynamic resampling filters groups with identical rewards, which provide no relative preference signal. Group Sequence Policy Optimization (GSPO)~\citep{zheng2025group} uses the length-normalized ratio $w_i(\theta)=[\pi_{\theta}(y_i\mid x_q)/\pi_{\theta_{\mathrm{old}}}(y_i\mid x_q)]^{1/|y_i|}$ and the clipped surrogate
\begin{equation}
\label{eq:gspo}
\mathcal{J}_{\mathrm{clip}}(\theta)=\mathbb{E}\Bigg[\frac{1}{G}\sum_{i=1}^{G}
\min\Big(w_i(\theta)\widehat{A}_i,\operatorname{clip}\big(w_i(\theta),1-\epsilon,1+\epsilon_{\mathrm{high}}\big)\widehat{A}_i\Big)\Bigg].
\end{equation}
The lower and upper clipping offsets are $\epsilon$ and $\epsilon_{\mathrm{high}}$. Training also applies Kullback-Leibler (KL) regularization to a reference policy, gradient-norm clipping, and learning-rate warm-up. Sequence-level optimization matches the granularity of rubric assessments, whose judgments about organization, support, and consistency concern the response as a whole. Because these properties depend on relations across the text, a dimension score need not be attributable to an individual token; the policy objective uses these response-level assessments directly.

\section{Experiments}
\label{sec:experiments}
The main benchmarks evaluate the complete writing pipeline. Training ablations compare inductive and direct rubric construction and the contribution of subsequent RL; transfer and evaluator studies examine rubric utility and the robustness of the resulting writers. Appendix~\ref{app:sensitivity} analyzes revision and decoding parameters.

\subsection{Experimental Setup}
\label{sec:setup}
\textbf{Benchmarks and metrics.}
WritingBench~\citep{wu2025writingbench} contains 1,000 requests spanning six primary domains and 100 subdomains, with style, format, and length constraints. We report scores from its WritingBench-Critic-Model-Qwen-7B evaluator. HelloBench~\citep{que2024hellobench} covers open-ended question answering, summarization, dialogue, text completion, and heuristic text generation. We use its evaluation program with DeepSeek-V3.2-Exp and report quality scores and mean word counts. We use the latest public benchmark releases available at evaluation time. Benchmark assessment uses their evaluation criteria, separately from the rubrics supplied to the writer.

\textbf{Baselines.}
Qwen3-4B and Qwen3-8B are the corresponding backbone references. Writing-specialized baselines include Suri-I-ORPO-7B, LongWriter-8B, WritingBench-7B, SuperWriter-7B, and LongWriter-Zero-32B. Larger general-purpose models contextualize the performance of the trained writers. Tables~\ref{main-performance1} and~\ref{main-performance2} list the evaluated checkpoints.

\textbf{Implementation.}
Writer SFT uses 15K instances and RL uses the remaining approximately 4K; both writer sizes use the same training hyperparameters. LLaMA-Factory~\citep{zheng2024llamafactory} implements SFT; MS-Swift~\citep{zhao2025swift} and DeepSpeed-ZeRO implement RL on NVIDIA A800 80GB GPUs. Qwen3-32B supplies training data and rubric-based rewards. For the main benchmarks, Writer-R1 receives RubricGen-4B criteria as additional input and generates a single response, using float16 precision, temperature 0.7, and top-$p$ 0.8. Appendix~\ref{app:optimization} gives training settings and reward implementation details.

\subsection{Main Results}
\label{sec:results}
\begin{table}[t]
\caption{Comparison of writing models on WritingBench. D1 to D6: Academic and Engineering, Finance and Business, Politics and Law, Literature and Art, Education, and Advertising and Marketing. R1 to R3: Style, Format, and Length; C1 to C3: corresponding category scores.}
\label{main-performance1}
\centering
\resizebox{0.97\textwidth}{!}{%
\begin{tabular}{lccccccccccccc}
\toprule
\multirow{2}{*}{\textbf{Models}} & \multirow{2}{*}{\textbf{Overall}} & \multicolumn{6}{c}{\textbf{Domains}} & \multicolumn{6}{c}{\textbf{Requirements}}    \\
\cmidrule(lr){3-8}
\cmidrule(lr){9-14}
 & & \textbf{D1} & \textbf{D2} & \textbf{D3} & \textbf{D4} &\textbf{D5} & \textbf{D6} & \textbf{R1} & \textbf{C1} &\textbf{R2} & \textbf{C2} & \textbf{R3} & \textbf{C3} \\
\midrule
\addlinespace[1pt]	
\rowcolor[rgb]{0.99,0.96,0.87} \multicolumn{14}{l}{\textbf{Mid-Scale General Models}} \\
\addlinespace[1pt]
\quad Mistral-Large-Instruct & 0.764 & 0.77 & 0.76 & 0.78 & 0.73 & 0.79 & 0.76 & 0.77 & 0.82 & 0.77 & 0.87 & 0.77 & 0.79 \\
\quad Llama-3.3-70B-Instruct & 0.701 & 0.70 & 0.69 & 0.70 & 0.68 & 0.73 & 0.73 & 0.71 & 0.78 & 0.71 & 0.82 & 0.70 & 0.72 \\
\quad Qwen-2.5-72B-Instruct & 0.743 & 0.77 & 0.74 & 0.76 & 0.69 & 0.78 & 0.73 & 0.75 & 0.79 & 0.76 & 0.86 & 0.74 & 0.75 \\
\quad Qwen3-32B & 0.788 & 0.78 & 0.77 & 0.79 & 0.79 & 0.80 & 0.81 & 0.80 & 0.84 & 0.79 & 0.86 & 0.79 & 0.80 \\
\midrule
\addlinespace[1pt]	
\rowcolor[rgb]{0.99,0.96,0.87} \multicolumn{14}{l}{\textbf{Large-Scale General Models}} \\
\addlinespace[1pt]	
\quad DeepSeek-R1 & 0.774 & 0.77 & 0.75 & 0.77 & 0.79 & 0.78 & 0.81 & 0.79 & 0.84 & 0.78 & 0.85 & 0.78 & 0.75 \\
\quad DeepSeek-R1-0528 & 0.829 & 0.83 & 0.82 & 0.82 & 0.85 & 0.82 & 0.84 & 0.84 & 0.86 & 0.83 & 0.89 & 0.84 & 0.84 \\
\quad DeepSeek-V3 & 0.745 & 0.74 & 0.73 & 0.74 & 0.76 & 0.74 & 0.78 & 0.77 & 0.82 & 0.75 & 0.84 & 0.74 & 0.72 \\
\quad DeepSeek-V3.1 & 0.795 & 0.80 & 0.79 & 0.78 & 0.82 & 0.79 & 0.80 & 0.80 & 0.84 & 0.80 & 0.88 & 0.80 & 0.81 \\
\quad DeepSeek-V3.2-Exp & 0.796 & 0.80 & 0.78 & 0.77 & 0.83 & 0.81 & 0.80 & 0.81 & 0.84 & 0.80 & 0.88 & 0.81 & 0.83 \\
\quad Qwen3-235B-A22B & 0.806 & 0.80 & 0.79 & 0.79 & 0.82 & 0.82 & 0.81 & 0.82 & 0.85 & 0.80 & 0.87 & 0.81 & 0.82 \\
\quad GLM-5 & 0.810 & 0.79 & 0.80 & 0.81 & 0.82 & 0.83 & 0.82 & 0.82 & 0.85 & 0.82 & 0.88 & 0.81 & 0.81 \\
\quad GLM-5.1 & 0.837 & 0.82 & 0.83 & 0.83 & 0.86 & 0.85 & 0.84 & 0.85 & 0.87 & 0.84 & 0.89 & 0.84 & 0.83 \\
\quad LongCat-Flash-Chat & 0.839 & 0.84 & 0.83 & 0.85 & 0.84 & 0.85 & 0.83 & 0.84 & 0.87 & 0.84 & 0.89 & 0.82 & 0.82 \\
\quad Kimi-K2 & 0.859 & 0.87 & 0.86 & 0.86 & 0.86 & 0.86 & 0.85 & 0.86 & 0.89 & 0.86 & 0.90 & 0.86 & 0.86 \\
\midrule
\addlinespace[1pt]	
\rowcolor[rgb]{0.99,0.96,0.87} \multicolumn{14}{l}{\textbf{Small-Scale or Task-Specific Models}} \\
\addlinespace[1pt]
\quad Qwen3-4B & 0.681 & 0.69 & 0.69 & 0.69 & 0.64 & 0.70 & 0.68 & 0.69 & 0.75 & 0.70 & 0.82 & 0.67 & 0.73 \\
\quad Qwen3-8B & 0.713 & 0.71 & 0.72 & 0.71 & 0.70 & 0.73 & 0.72 & 0.72 & 0.78 & 0.72 & 0.84 & 0.72 & 0.77 \\
\quad Suri-I-ORPO-7B & 0.497 & 0.56 & 0.53 & 0.50 & 0.41 & 0.50 & 0.51 & 0.48 & 0.52 & 0.50 & 0.54 & 0.45 & 0.40 \\
\quad LongWriter-8B & 0.627 & 0.64 & 0.64 & 0.63 & 0.60 & 0.65 & 0.60 & 0.63 & 0.74 & 0.63 & 0.67 & 0.63 & 0.68 \\
\quad WritingBench-7B & 0.744 & 0.74 & 0.72 & 0.75 & 0.73 & 0.77 & 0.77 & 0.75 & 0.84 & 0.76 & 0.81 & 0.74 & 0.72 \\
\quad SuperWriter-7B & 0.756 & 0.75 & 0.75 & 0.75 & 0.75 & 0.79 & 0.77 & 0.77 & 0.81 & 0.76 & 0.84 & 0.76 & 0.73 \\
\quad LongWriter-Zero-32B & 0.802 & 0.81 & 0.80 & 0.80 & 0.79 & 0.83 & 0.81 & 0.81 & 0.84 & 0.81 & 0.87 & 0.80 & 0.83 \\
\rowcolor[rgb]{0.91,0.96,0.91} \quad  Writer-R1-4B & 0.863 & 0.87 & 0.88 & 0.86 & 0.84 & 0.88 & 0.86 & 0.86 & 0.88 & 0.87 & 0.92 & 0.86 & 0.86 \\
\rowcolor[rgb]{0.91,0.96,0.91} \quad  Writer-R1-8B & 0.871 & 0.87 & 0.88 & 0.87 & 0.88 & 0.88 & 0.87 & 0.87 & 0.90 & 0.87 & 0.91 & 0.86 & 0.85 \\
\bottomrule
\end{tabular}%
}
\end{table}

\textbf{General writing quality and requirement satisfaction.}
Writer-R1-4B and Writer-R1-8B attain the two highest overall WritingBench scores, 0.863 and 0.871 (Table~\ref{main-performance1}), improving over their backbones by 0.182 and 0.158. Both exceed LongWriter-Zero-32B at 0.802 and Kimi-K2 at 0.859, the strongest writing-specialized and general-purpose baselines in this comparison. Improvements cover all six domains and all reported requirement metrics. In Literature and Art, the 4B model rises from 0.64 to 0.84 and the 8B model from 0.70 to 0.88. These comparisons support the effectiveness of the complete pipeline across the evaluated writing tasks; the training ablations examine how performance changes with rubric construction and training stage.

The requirement-level results distinguish the strengths of the two model scales. Writer-R1-8B scores higher on the style category, 0.90 versus 0.88, while Writer-R1-4B scores slightly higher on format and length categories, 0.92 versus 0.91 and 0.86 versus 0.85. The larger writer attains a higher overall score, although the advantage does not extend uniformly to every requirement metric. The shared gains over the backbones are broader than these differences between the trained variants.

\begin{table}[t]
\caption{Comparison of response quality and mean word counts on HelloBench. OEQA: open-ended question answering; Summ: summarization; Chat: dialogue; TC: text completion; HTG: heuristic text generation.}
\label{main-performance2}
\centering
\resizebox{1\textwidth}{!}{%
\begin{tabular}{lcccccccccccc}
\toprule
\multirow{2}{*}{\textbf{Models}} & \multicolumn{2}{c}{\textbf{Overall}} & \multicolumn{2}{c}{\textbf{OEQA}}& \multicolumn{2}{c}{\textbf{Summ}}& \multicolumn{2}{c}{\textbf{Chat}}& \multicolumn{2}{c}{\textbf{TC}}& \multicolumn{2}{c}{\textbf{HTG}}    \\	 		 		 		 	
\cmidrule(lr){2-3}\cmidrule(lr){4-5}\cmidrule(lr){6-7}\cmidrule(lr){8-9}\cmidrule(lr){10-11}\cmidrule(lr){12-13}
 & \textbf{Score} & \textbf{Length} & \textbf{Score} & \textbf{Length} &\textbf{Score} & \textbf{Length} & \textbf{Score} & \textbf{Length} &\textbf{Score} & \textbf{Length} & \textbf{Score} & \textbf{Length} \\
\midrule
Qwen3-4B & 0.641 & 1,626.61 & 0.547 & 1,634.09 & 0.736 & 659.04 & 0.646 & 2,380.94 & 0.616 & 1,992.83 & 0.662 & 1,466.15 \\
Qwen3-8B & 0.682 & 1,561.45 & 0.583 & 1,671.10 & 0.743 & 608.35 & 0.725 & 2,364.53 & 0.649 & 1,910.64 & 0.709 & 1,252.61 \\
WritingBench-7B & 0.636 & 1,012.63 & 0.553 & 924.07 & 0.631 & 561.09 & 0.619 & 1,376.21 & 0.655 & 1,174.14 & 0.722 & 1,027.65 \\
SuperWriter-7B & 0.579 & 9,497.69 & 0.478 & 9,423.32 & 0.517 & 9,078.57 & 0.571 & 9,563.18 & 0.627 & 9,111.82 & 0.700 & 10,311.54 \\
LongWriter-Zero-32B & 0.713 & 2,711.94 & 0.769 & 2,004.73 & 0.672 & 935.03 & 0.690 & 3,443.42 & 0.659 & 4,935.58 & 0.776 & 2,240.93 \\
\rowcolor[rgb]{0.91,0.96,0.91}Writer-R1-4B & 0.700 & 1,997.75 & 0.625 & 1,939.67 & 0.685 & 619.81 & 0.685 & 1,945.70 & 0.714 & 3,755.66 & 0.792 & 1,727.88 \\
\rowcolor[rgb]{0.91,0.96,0.91}Writer-R1-8B & 0.733 & 1,835.00 & 0.670 & 1,790.25 & 0.756 & 629.46 & 0.717 & 1,999.16 & 0.725 & 2,921.39 & 0.798 & 1,834.73 \\
\bottomrule
\end{tabular}%
}
\end{table}

\textbf{Long-form quality and output length.}
On HelloBench (Table~\ref{main-performance2}), Writer-R1-4B improves from 0.641 to 0.700 and Writer-R1-8B from 0.682 to 0.733. The 8B model leads the evaluated systems overall, exceeding LongWriter-Zero-32B by 0.020, and ranks first on summarization, text completion, and heuristic text generation. Both variants improve on four of five tasks; summarization decreases for the 4B model and dialogue decreases slightly for the 8B model. Writer-R1-8B achieves its higher overall score with a mean of 1,835 words, compared with 2,712 for LongWriter-Zero-32B and 9,498 for SuperWriter-7B. Thus, its quality advantage coexists with substantially shorter outputs. The task breakdown also indicates that the pipeline does not improve every form of long-text generation equally: LongWriter-Zero-32B retains the strongest open-ended question answering score. Reporting quality alongside length helps distinguish the observed improvements from a simple tendency to produce longer responses.

\subsection{Training Ablations: Rubric Construction and RL}
\label{sec:ablations}
We use \textbf{Direct Rubric Generation (DRG)} as a comparison baseline: Qwen3-32B generates a complete prompt-specific rubric in one call, without the separate discovery and consolidation operations of IRG. Table~\ref{tab:training-ablation} compares the backbone, SFT without rubric conditioning, and rubric-conditioned training with either construction method. For each construction method, we build its rubric corpus, select demonstrations from five Qwen3-32B candidates, and train the associated writer and rubric generator. The downstream training procedures and budgets are matched, while the generated supervision follows the corresponding rubric construction method. Each RL variant continues from its associated SFT model and uses that method for reward criteria. Inference uses the generator trained on the corresponding rubric corpus. The comparison evaluates each construction procedure throughout the pipeline, including the supervision and inference guidance it produces.

\begin{table}[t]
\caption{WritingBench training ablations with Qwen3-4B; IRG + SFT + RL is Writer-R1-4B. Bold marks column maxima. D1 to D6: Academic and Engineering, Finance and Business, Politics and Law, Literature and Art, Education, and Advertising and Marketing.}
\label{tab:training-ablation}
\centering
\small
\setlength{\tabcolsep}{5pt}
\begin{tabular}{@{}lccccccc@{}}
\toprule
\textbf{Training configuration} & \textbf{Overall} & \textbf{D1} & \textbf{D2} & \textbf{D3} & \textbf{D4} & \textbf{D5} & \textbf{D6} \\
\midrule
Qwen3-4B & 0.681 & 0.69 & 0.69 & 0.69 & 0.64 & 0.70 & 0.68 \\
SFT & 0.725 & 0.77 & 0.72 & 0.72 & 0.66 & 0.77 & 0.73 \\
\addlinespace[2pt]
DRG + SFT & 0.719 & 0.78 & 0.72 & 0.71 & 0.65 & 0.76 & 0.72 \\
DRG + SFT + RL & 0.766 & 0.74 & 0.74 & 0.79 & 0.78 & 0.77 & 0.79 \\
\addlinespace[2pt]
IRG + SFT & 0.816 & 0.85 & 0.82 & 0.81 & 0.77 & 0.84 & 0.81 \\
\rowcolor{lightg} IRG + SFT + RL & \textbf{0.863} & \textbf{0.87} & \textbf{0.88} & \textbf{0.86} & \textbf{0.84} & \textbf{0.88} & \textbf{0.86} \\
\bottomrule
\end{tabular}
\end{table}

\textbf{Inductive construction yields stronger writing models.}
IRG exceeds DRG by 0.097 after SFT and by the same margin after SFT+RL, with higher scores in all six domains at both stages. IRG+SFT also exceeds SFT without rubrics by 0.091, whereas DRG+SFT scores 0.006 lower than SFT alone. The difference between these variants indicates that the utility of rubric conditioning depends on how the specifications are constructed. IRG+SFT also outperforms DRG+SFT+RL by 0.050. The advantage is therefore present before RL and persists even when the direct-generation pipeline receives the additional optimization stage, supporting the utility of IRG within the complete training and inference workflow.

\textbf{RL improves both rubric-based variants.}
Subsequent RL increases the overall score by 0.047 under both DRG and IRG, reaching 0.766 and 0.863. The equal absolute gains preserve the performance gap between the two construction methods. At the domain level, IRG+SFT+RL improves over IRG+SFT throughout, whereas DRG+SFT+RL decreases from 0.78 to 0.74 in Academic and Engineering despite its overall gain. The aggregate improvement from RL thus coexists with domain-specific variation, and IRG retains the stronger result in every reported domain. Together, these comparisons support two contributions to the final pipeline: the stronger performance associated with inductively constructed rubrics and the additional improvement from optimizing responses against the corresponding criteria.

\subsection{Rubric Transfer Across Writing Models}
\label{sec:analysis}
\begin{figure}[t]
    \centering
    \includegraphics[width=0.65\textwidth]{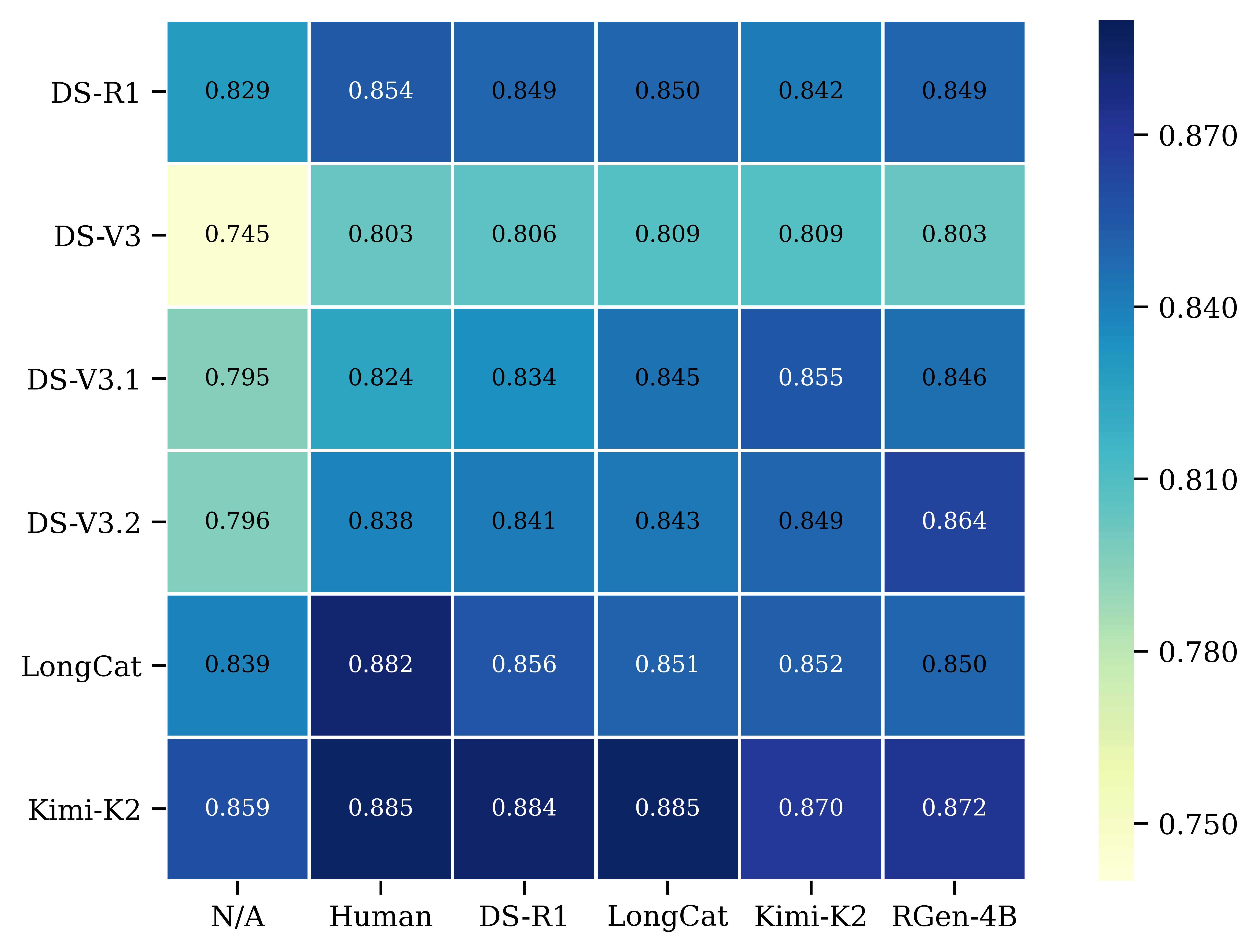}
    \caption{Rubric transfer on WritingBench across writers (rows) and rubric sources (columns), with writer parameters fixed. DS-R1: DeepSeek-R1-0528; DS-V3: DeepSeek-V3; DS-V3.1: DeepSeek-V3.1; DS-V3.2: DeepSeek-V3.2-Exp; LongCat: LongCat-Flash-Chat; RGen-4B: RubricGen-4B. N/A: direct answering without feedback; Human: human-curated rubrics.}
    \label{fig:rubric-transfer}
\end{figure}

We assess rubric utility with fixed target writers by varying the criteria supplied as feedback on their outputs. Human-curated and automatically generated rubrics are applied within the same feedback protocol, and the resulting texts are scored with the WritingBench critic. Figure~\ref{fig:rubric-transfer} compares rubric sources across six target writers.

\textbf{Cross-model rubric transfer.}
Kimi-K2 rubrics yield scores of 0.809, 0.855, and 0.849 on DeepSeek-V3, DeepSeek-V3.1, and DeepSeek-V3.2-Exp, exceeding the human-curated references of 0.803, 0.824, and 0.838. These comparisons show that generated criteria can guide writers other than their constructor. Utility nevertheless depends on the target: human-curated criteria lead for LongCat-Flash-Chat, whereas RubricGen-4B leads for DeepSeek-V3.2-Exp. This study assesses rubric-guided revision with fixed writer parameters, complementing the training comparisons.

\textbf{Compact rubric generation.}
RubricGen-4B learns from all 19K mined prompt and rubric pairs. Its criteria match the human reference on DeepSeek-V3 and exceed it on DeepSeek-V3.1 and DeepSeek-V3.2-Exp. With DeepSeek-V3.2-Exp, the revised responses score 0.864, compared with 0.838 under human-curated criteria and 0.849 under Kimi-K2 criteria. The results support transferring the mined supervision to a compact generator, which also supplies the criteria used in Writer-R1 inference. Together with the training ablations, these results support two uses of the corpus: learning a writer that follows explicit requirements and learning a generator that expresses those requirements for other writers. Appendix~\ref{app:rubric-use} gives the feedback protocol and detailed source comparisons.

\subsection{Robustness Across LLM and Human Evaluators}
\label{sec:evaluators}
\begin{figure}[t]
    \centering
    \includegraphics[width=0.83\textwidth]{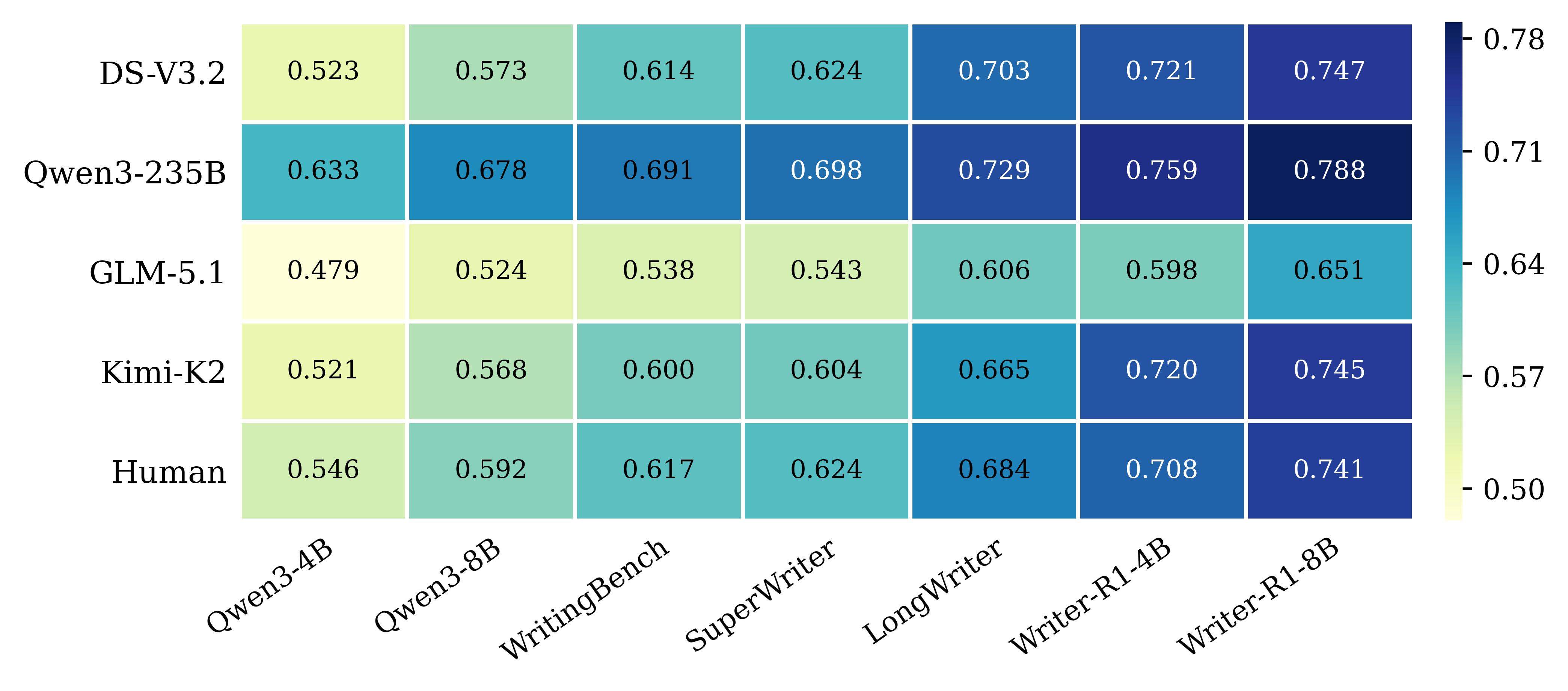}
    \caption{WritingBench evaluation by LLMs and humans (rows), comparing writers (columns) within rows. Human scores average four expert assessments. DS-V3.2: DeepSeek-V3.2-Exp; Qwen3-235B: Qwen3-235B-A22B; WritingBench: WritingBench-7B; SuperWriter: SuperWriter-7B; LongWriter: LongWriter-Zero-32B.}
    \label{fig:evaluator-transfer}
\end{figure}

We test sensitivity to the final evaluation source using DeepSeek-V3.2-Exp, Qwen3-235B-A22B, GLM-5.1, Kimi-K2, and human assessment. Four experts independently assess the anonymized responses using the same criteria, scoring scale, and aggregation rules as the LLM evaluation; reported human scores average their four assessments. Figure~\ref{fig:evaluator-transfer} reports the overall scores. Appendix~\ref{app:evaluation} details the evaluation protocol and gives domain and requirement scores for three LLM evaluators.

Writer-R1-8B ranks first overall under every evaluation source, and both variants outperform their backbones under all four LLM evaluators and human assessment. Human scores rise from 0.546 to 0.708 at the 4B scale and from 0.592 to 0.741 at the 8B scale, exceeding LongWriter-Zero-32B at 0.684. Some pairwise rankings vary: GLM-5.1 places Writer-R1-4B slightly below LongWriter-Zero-32B, whereas the other sources rank it higher. The consistent backbone improvements and the lead of Writer-R1-8B support the robustness of the pipeline across assessment sources.

\section{Conclusion}
\label{sec:conclusion}
We presented Writer-R1, which uses IRG to turn writing requirements into task-specific rubrics for both model conditioning and RL rewards. This shared specification makes assessment criteria available during generation and connects rubric construction with policy learning. Writer-R1 improves over its backbones on WritingBench and HelloBench; on HelloBench, the 8B model achieves 0.733 compared with 0.682 for its backbone. Training comparisons favor IRG over direct generation. Rubric transfer and compact generation support the reuse of mined specifications, while LLM and human evaluations establish consistent gains across assessment sources. These complementary results support inductive rubric construction as a practical basis for learning open-ended writing.

\clearpage
\bibliographystyle{iclr2027_conference}
\bibliography{iclr2027_conference}

\appendix

\section{Corpus Construction and Rubric Mining Details}
\label{app:mining}
This appendix details corpus construction, rubric representation, and output validation. Algorithm~\ref{alg:writer-r1} summarizes construction, training, and inference. Appendix~\ref{app:prompts} collects the operational prompts.

\subsection{Prompt Expansion, Quality Review, and Corpus Reuse}
\label{app:data}
We expand the 6K LongWrite training prompts~\citep{wu2025longwriter} using Qwen3-32B. The purpose is to increase the variety of writing requests available for training while preserving meaningful content, context length, and task difficulty. Reconstruction attempts are allocated in a 1:1 ratio to the following strategies.

\textbf{Length-aware reconstruction.}
The rewriting prompt requests a new instruction that preserves the high-level task, language, and difficulty while substantially changing wording, organization, scenario details, and any supplied source material. It asks the model to match the original character length. This is a generation instruction rather than an exact-length decoding constraint. The implementation records original and reconstructed character lengths and excludes source inputs exceeding 8,000 whitespace-delimited segments.

\textbf{Reconstruction with length recovery.}
The second strategy starts from the same rewrite, then expands reconstructions that are too short. It uses the number of whitespace-delimited segments as a length proxy. If the new count is no greater than three quarters of the source count, the model receives an enrichment instruction; up to six additional calls are permitted. For material-based tasks, enrichment develops a fuller, coherent replacement passage. Otherwise, it clarifies the existing task without introducing unrelated requirements or a large body of independent material. This procedure preserves long-context training opportunities that an unconstrained rewrite may shorten.

\textbf{Quality filtering and manual review.}
Automatic filtering removes outputs with consecutively repeated string fragments, which can arise during long-context generation. Manual review assesses whether the reconstruction remains a reasonable, usable writing instruction. The review focuses on:
\begin{enumerate}
\setlength{\itemsep}{2pt}
    \item \textbf{Task clarity.} The requested writing product and its purpose are identifiable; the output is an instruction rather than an answer, editing commentary, or an incomplete fragment.
    \item \textbf{Intent and constraint consistency.} The reconstruction preserves the broad task category and comparable difficulty. Requirements on language, style, structure, and length are mutually compatible.
    \item \textbf{Contextual coherence.} Supplied materials, entities, events, and references form a coherent context. Instructions referring to a passage or other provided material include the material needed to carry out the task.
    \item \textbf{Substantive variation.} Changes extend beyond minor lexical substitutions and introduce meaningful variation in content or setting without redirecting the task toward an unrelated objective.
    \item \textbf{Textual integrity and usable length.} The instruction and any embedded material remain readable and complete, with sufficient detail for the intended task and no degenerate repetition or abrupt truncation.
\end{enumerate}
These criteria organize the assessment of prompt usability and diversity. They do not impose a preferred writing answer or require a particular rhetorical approach.

\textbf{Screening against evaluation prompts.}
We tokenize each training candidate and benchmark prompt with the Qwen3-4B tokenizer. For a token sequence $T(q)=(t_1,\ldots,t_m)$, let $\mathcal{G}_3(q)=\{(t_i,t_{i+1},t_{i+2}):1\leq i\leq m-2\}$ be its set of contiguous token trigrams. We apply the same construction to each benchmark sample $e$ and compute
\begin{equation}
\label{eq:overlap}
O(q,e)=\frac{|\mathcal{G}_3(q)\cap\mathcal{G}_3(e)|}{|\mathcal{G}_3(e)|},
\qquad
q\ \text{is discarded if}\ \max_{e\in\mathcal{T}}O(q,e)\geq0.10,
\end{equation}
where $\mathcal{T}$ contains the WritingBench and HelloBench evaluation prompts. The denominator is the compared test sample, so the measure captures the proportion of its distinct token trigrams shared with the training candidate. Very few candidates reach this threshold. The same tokenizer and overlap rule are used for both reconstruction strategies and both benchmarks.

After filtering and review, the corpus contains 19K instances. Qwen3-32B mines a rubric for each final prompt, so the criteria reflect its reconstructed content and constraints. Writer SFT uses 15K instances with the highest-scoring response among five Qwen3-32B candidates as the target; the remaining approximately 4K instances support RL. During both stages, the writer input directly concatenates the prompt with the serialized rubric. RubricGen-4B uses all 19K prompt and rubric pairs to learn the separate mapping from writing requests to criteria.

Constructing rubrics after prompt reconstruction keeps their scopes consistent with the final training requests. Each SFT rubric guides demonstration selection and writer conditioning; each RL rubric supplies the conditioning context and assessment criteria for policy samples.

\subsection{Candidate Dimensions and Consolidation}
Requirement discovery targets 15 to 20 candidate dimensions and produces $\mathcal{D}^{(0)}(q)=\{(d_j,a_j)\}_{j=1}^{N(q)}$, linking candidate dimensions to task objectives, genre conventions, audience expectations, and explicit constraints. Discovery seeks broad coverage while discouraging synonymous repetition; consolidation subsequently reviews the proposals for residual redundancy and differences in granularity.

Dimension consolidation targets 5 to 8 retained dimensions and maps these proposals to $\mathcal{D}^{(1)}(q)$ by merging synonymous dimensions, clarifying overlapping assessment targets, and discarding marginal criteria. Retention depends on relevance, observability, and the information contributed beyond the remaining dimensions. The retained scope description $a_k$ states what should be assessed. Content coverage, for example, concerns whether required material is present, whereas organization concerns how that material is arranged. The dimensions need not be statistically independent; the purpose of consolidation is to prevent repeated assessment of the same target from unintentionally dominating the reward.

\subsection{Scoring Intervals and Performance Descriptions}
For each retained dimension, scoring specification constructs
\begin{equation}
\label{eq:bins}
\mathcal{B}_k=\{(I_{k,m},\rho_{k,m})\}_{m=1}^{M_k},
\qquad I_{k,m}=\{2m-1,2m\},
\qquad s_k^{\max}=2M_k.
\end{equation}
Here, each interval contains two consecutive integer scores, $M_k$ is the number of intervals, and $\rho_{k,m}$ describes the associated performance level. Thus, the intervals cover all permitted integer scores from 1 to $2M_k$ without overlap. Variable interval counts give different dimension maxima. In \eqref{eq:reward}, these maxima determine the weights of normalized dimension scores as well as the aggregate denominator.

The instructions request 5 to 8 score intervals and progressively stronger satisfaction of the same assessment scope. Distinctions should rest on observable properties, such as omitted requirements, unsupported claims, imprecise interpretation, or compliance with the requested structure. Adjacent descriptions should identify differences an evaluator can recognize in the response. These specifications guide construction of the criteria. Their downstream utility is evaluated through the training comparisons in Section~\ref{sec:ablations}.

\begin{algorithm}[t]
\caption{Writer-R1 corpus construction, training, and inference}
\label{alg:writer-r1}
\begin{algorithmic}[1]
\REQUIRE Expanded LongWrite prompts $\mathcal{D}$; writer $\pi_\theta$; generator $f_\phi$; group size $G$; Qwen3-32B constructor and evaluator $\mathcal{E}$
\ENSURE Trained writer $\pi_\theta$ and rubric generator $f_\phi$
\STATE Initialize rubric corpus $\mathcal{B}\leftarrow\emptyset$.
\FOR{each prompt $q\in\mathcal{D}$}
    \STATE Discover dimensions $\mathcal{D}^{(0)}\leftarrow A_{\mathrm{disc}}(q)$.
    \STATE Consolidate dimensions $\mathcal{D}^{(1)}\leftarrow A_{\mathrm{cons}}(q,\mathcal{D}^{(0)})$.
    \STATE Construct rubric $\mathcal{R}(q)\leftarrow A_{\mathrm{spec}}(q,\mathcal{D}^{(1)})$.
    \STATE Store $(q,\mathcal{R}(q))$ in $\mathcal{B}$.
\ENDFOR
\STATE Allocate 15K records to writer SFT and the remainder to $\mathcal{B}_{\mathrm{RL}}$.
\STATE For each SFT record, generate five Qwen3-32B candidates and retain the highest-scoring response under its rubric.
\STATE Train $\pi_\theta$ by \eqref{eq:sft-loss} with $x_q=q\oplus\operatorname{serialize}(\mathcal{R}(q))$.
\STATE Train $f_\phi$ on all 19K prompt and rubric pairs.
\FOR{each RL training step}
    \STATE Retrieve $(q,\mathcal{R}(q))$ from $\mathcal{B}_{\mathrm{RL}}$; set $x_q=q\oplus\operatorname{serialize}(\mathcal{R}(q))$.
    \STATE Sample $\{y_i\}_{i=1}^{G}\sim\pi_{\theta_{\mathrm{old}}}(\cdot\mid x_q)$.
    \FOR{$i=1$ to $G$}
        \STATE Score $y_i$ with $\mathcal{E}$ using $\mathcal{R}(q)$ and aggregate by \eqref{eq:reward}.
    \ENDFOR
    \STATE Resample degenerate groups and compute advantages by \eqref{eq:advantage}.
    \STATE Update $\pi_\theta$ with the GSPO surrogate and KL regularization.
    \STATE Refresh the behavior policy according to the sampling schedule.
\ENDFOR
\STATE For a new request $q$, compute $\widehat{\mathcal{R}}(q)=f_\phi(q)$.
\STATE Generate $y\sim\pi_\theta(\cdot\mid q\oplus\operatorname{serialize}(\widehat{\mathcal{R}}(q)))$.
\end{algorithmic}
\end{algorithm}

\subsection{Output Parsing and Rubric Assembly}
\label{app:rubric-parsing}
The implementation parses \texttt{name} and \texttt{reason} records and removes repeated dimension names while preserving order. The parser applies acceptance bounds separately from the generation targets: discovery requires at least two parsed candidates, consolidation requires at least four dimensions and truncates excess output to eight, and each scoring specification requires at least four parsed intervals. Descriptions are serialized with consecutive two-score labels. A completed rubric must contain at least three successfully scored dimensions; otherwise, the record is excluded. Final dimensions retain their consolidation order. These checks enforce a usable structure before the rubric is stored and reused.

\section{Optimization and Implementation Details}
\label{app:optimization}
\subsection{Supervised Training of Writers and the Rubric Generator}
Writer SFT uses the conditional loss in \eqref{eq:sft-loss}. Let $\mathcal{C}_5(q)$ contain five Qwen3-32B responses to the request. The supervised target is
\begin{equation}
\label{eq:sft-selection}
y^\star(q)=\operatorname*{arg\,max}_{y\in\mathcal{C}_5(q)}r(q,y).
\end{equation}
All five candidates are evaluated against the associated rubric, and the selected response is paired with $x_q$, the concatenated request and rubric. Writer SFT uses 15K such demonstrations before RL on the remaining approximately 4K instances. This allocation separates the prompt subsets while preserving their input representation and rubric-based assessment.

RubricGen-4B uses the same autoregressive training objective with the writing request as input and $\operatorname{serialize}(\mathcal{R}(q))$ as the target, using all 19K prompt and rubric pairs. Its targets contain the mined dimensions, assessment scopes, and score descriptions. The writer learns to produce a response conditioned on a rubric, whereas the generator learns to produce the rubric from the request.

\subsection{Group Sequence Policy Optimization}
The probability of a response factorizes conditional on both the prompt and rubric:
\begin{equation}
\pi_{\theta}(y_i\mid x_q)=\prod_{t=1}^{|y_i|}\pi_{\theta}(y_{i,t}\mid x_q,y_{i,<t}).
\end{equation}
Substituting this factorization into the GSPO ratio~\citep{zheng2025group} gives
\begin{equation}
w_i(\theta)=\exp\!\left(\frac{1}{|y_i|}\sum_{t=1}^{|y_i|}
\log\frac{\pi_{\theta}(y_{i,t}\mid x_q,y_{i,<t})}
{\pi_{\theta_{\mathrm{old}}}(y_{i,t}\mid x_q,y_{i,<t})}\right).
\end{equation}
This geometric mean of token probability ratios is applied to the response-level advantage. In \eqref{eq:gspo}, the expectation is over records in the RL portion of the corpus and response groups sampled from the behavior policy using their combined contexts. Training maximizes the clipped surrogate with a KL penalty, written as $\mathcal{J}_{\mathrm{train}}=\mathcal{J}_{\mathrm{clip}}-\beta\mathcal{L}_{\mathrm{KL}}$, where $\mathcal{L}_{\mathrm{KL}}$ regularizes deviations from a reference policy and $\beta=0.01$.

Groups contain eight candidate responses. Dynamic sampling resamples groups without reward variation, with up to three resampling attempts. The lower and upper clipping offsets are $3\times10^{-4}$ and $4\times10^{-4}$. Rollout generation uses temperature 1.0, top-$p$ 0.9, top-$k$ 50, and a maximum response length of 4,096 tokens. These sampling settings govern RL exploration and differ from the decoding settings used for benchmark responses.

\subsection{Implementation and Training Settings}
The writing backbones are Qwen3-4B and Qwen3-8B. Qwen3-32B generates the training data and evaluates RL responses using their stored rubrics. RubricGen-4B supplies inference-time criteria. Final WritingBench scoring uses its dedicated critic, while HelloBench uses DeepSeek-V3.2-Exp. Keeping these roles explicit distinguishes the construction of writing specifications, their use for training rewards, and benchmark assessment.

Table~\ref{tab:training-settings} summarizes the shared 4B and 8B writer training hyperparameters. SFT uses LLaMA-Factory, and GSPO uses the sequence-level importance-sampling implementation in MS-Swift. Both stages use full-parameter training with bfloat16 precision and DeepSpeed ZeRO stage 3. GSPO uses four training processes with four gradient-accumulation steps, an input cutoff of 8,192 tokens, and a 4,096-token completion limit. Inputs exceeding the cutoff are truncated from the left. Evaluation and checkpointing occur every 20 training steps. Training runs on NVIDIA A800 80GB GPUs. Benchmark generation uses float16, temperature 0.7, and top-$p$ 0.8.

\begin{table}[t]
\caption{Shared training hyperparameters for Writer-R1-4B and Writer-R1-8B. The prompt and rubric form the writer input in both stages.}
\label{tab:training-settings}
\centering
\small
\begin{tabular}{@{}lll@{}}
\toprule
\textbf{Stage} & \textbf{Setting} & \textbf{Value} \\
\midrule
SFT & Allocated instances & 15K \\
 & Teacher candidates per prompt & 5 \\
 & Learning rate & $1\times10^{-5}$ \\
 & Epochs & 1 \\
 & Learning-rate schedule & Cosine \\
 & Warm-up fraction & 0.10 \\
 & Sequence cutoff & 16,384 tokens \\
 & Microbatch per device & 1 \\
 & Gradient accumulation & 16 steps \\
 & Validation fraction & 0.002 \\
\midrule
GSPO & Allocated instances & Approximately 4K \\
 & Learning rate & $1\times10^{-6}$ \\
 & Epochs & 2 \\
 & Warm-up fraction & 0.01 \\
 & Maximum input length & 8,192 tokens \\
 & Gradient accumulation & 4 steps \\
 & Training processes & 4 \\
 & Responses per prompt & 8 \\
 & Maximum resampling attempts & 3 \\
 & Maximum response length & 4,096 tokens \\
 & Lower and upper clipping offsets & $3\times10^{-4}$, $4\times10^{-4}$ \\
 & KL coefficient & 0.01 \\
 & Maximum gradient norm & 1.0 \\
 & Microbatch per device & 1 \\
 & Validation fraction & 0.05 \\
 & Rollout temperature, top-$p$, top-$k$ & 1.0, 0.9, 50 \\
\midrule
RubricGen-4B & Prompt and rubric pairs & 19K \\
\bottomrule
\end{tabular}
\end{table}

\subsection{Direct-Generation Baseline Protocol}
\label{app:ablation-protocol}
The DRG baseline in Section~\ref{sec:ablations} replaces IRG with a single Qwen3-32B call that directly constructs a complete rubric for the writing request. Its rubrics are then used for best-of-five demonstration selection, writer conditioning, and reward computation under the same downstream procedures and training budgets. Each construction method produces its own rubric corpus and selected demonstrations. A separate compact generator is trained on each rubric corpus and supplies criteria to the associated writer at inference. The generated supervision and inference criteria therefore vary with the construction method. This design evaluates the practical effect of using the complete induction procedure throughout writing training and deployment.

\subsection{Reward Evaluation and Aggregation}
The reward evaluator receives the candidate response and its serialized rubric. Each item is assessed in rubric order, returning the dimension name, an integer score with its maximum, and comments describing strengths and possible improvements. The complete scoring instruction is given in Appendix~\ref{app:reward-prompt}.

The implementation parses the item scores and maxima, discarding entries with a nonpositive maximum, a negative score, or a score exceeding its maximum. Valid scores are summed and divided by the sum of their maxima, then constrained to $[0,1]$. The numerical ratio supplies the RL reward; textual comments are used by the auxiliary revision protocol. A failed scoring call or an output with no valid scores receives the implementation fallback reward of 0.4. This fallback handles scoring failures and is separate from the assessment of response quality.

\section{Rubric Utility, Transfer, and Compact Generation}
\label{app:rubric-use}
The training ablations evaluate IRG within the complete writing pipeline. The following studies examine whether mined rubrics also guide revision with fixed writer parameters. They evaluate rubric sources through the quality of the revised texts, including criteria supplied by the compact generator trained on the mined corpus. Table~\ref{criteria_source1} provides detailed comparisons for three writers, and Figure~\ref{fig:rubric-transfer} covers six.

\subsection{Rubric-Guided Revision at Inference Time}
\label{app:feedback}
\begin{figure}[t]
    \centering
    \includegraphics[width=0.80\textwidth]{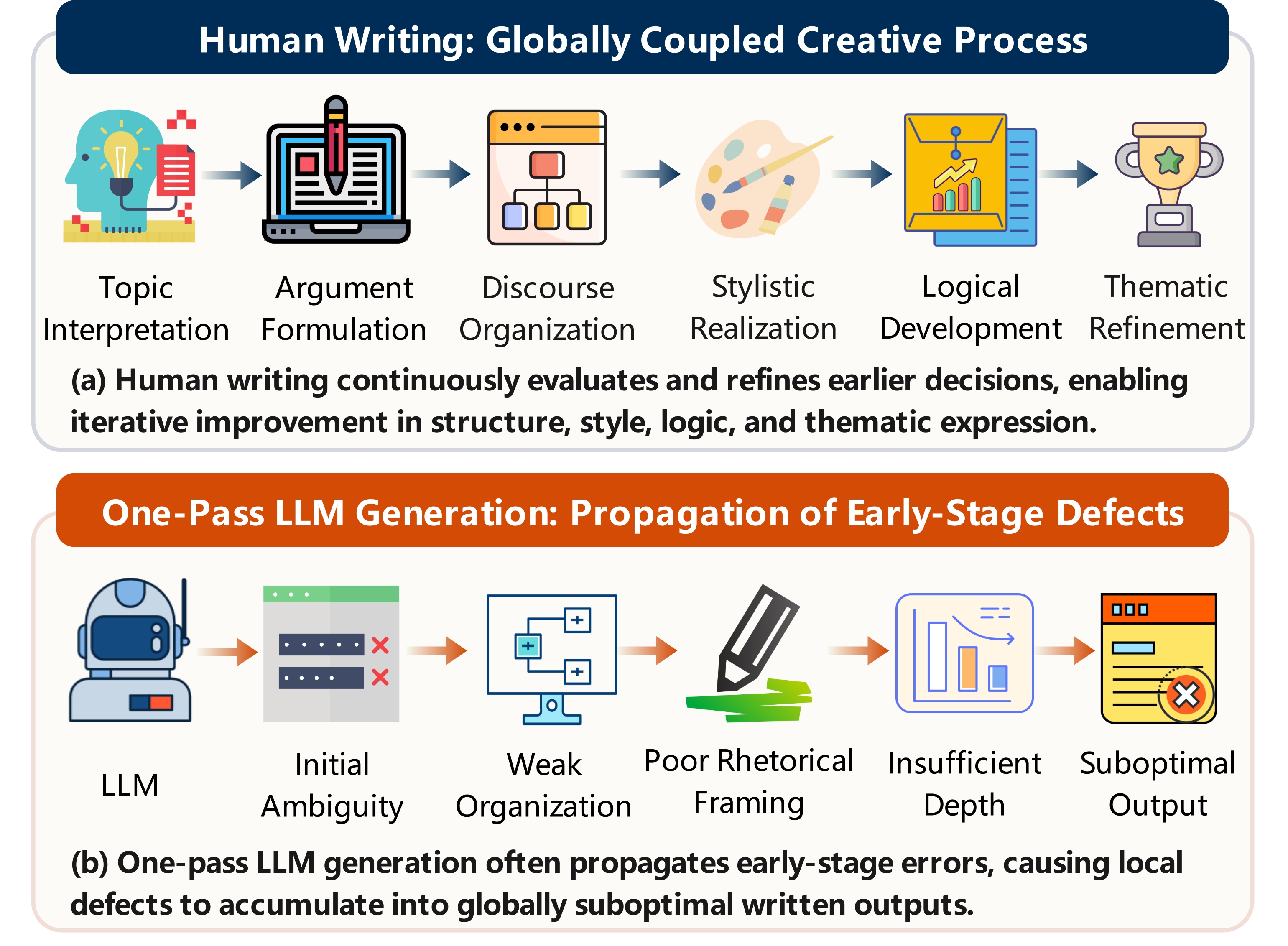}
    \caption{Illustration of interdependent writing decisions and their implications for draft assessment. (a) Feedback can inform related decisions about topic interpretation, argument, organization, style, logic, and theme. (b) Defects in early decisions can affect later parts of a response. Explicit criteria provide assessment targets for revision.}
    \label{fig:writing-motivation}
\end{figure}

Figure~\ref{fig:writing-motivation} illustrates how decisions about content, organization, and style can interact across a response. Rubric-guided revision uses explicit criteria to assess these properties in a completed draft.

The target writer first generates a draft from the writing request. Rubric-based assessment identifies unmet requirements, and the writer then revises the draft using the original request, previous response, and review. The default protocol permits five generation-and-assessment rounds, including the initial draft, and therefore at most four revisions. It accepts a response when its normalized rubric score reaches 0.9. If the budget is exhausted, it returns the highest-scoring draft encountered. Model parameters remain fixed throughout. This supplementary protocol, related to verbal self-reflection~\citep{shinn2023reflexion}, tests whether the criteria provide actionable feedback beyond their use in Writer-R1 training and rubric-conditioned response generation.

\begin{figure}[t]
    \centering
    \includegraphics[width=\textwidth]{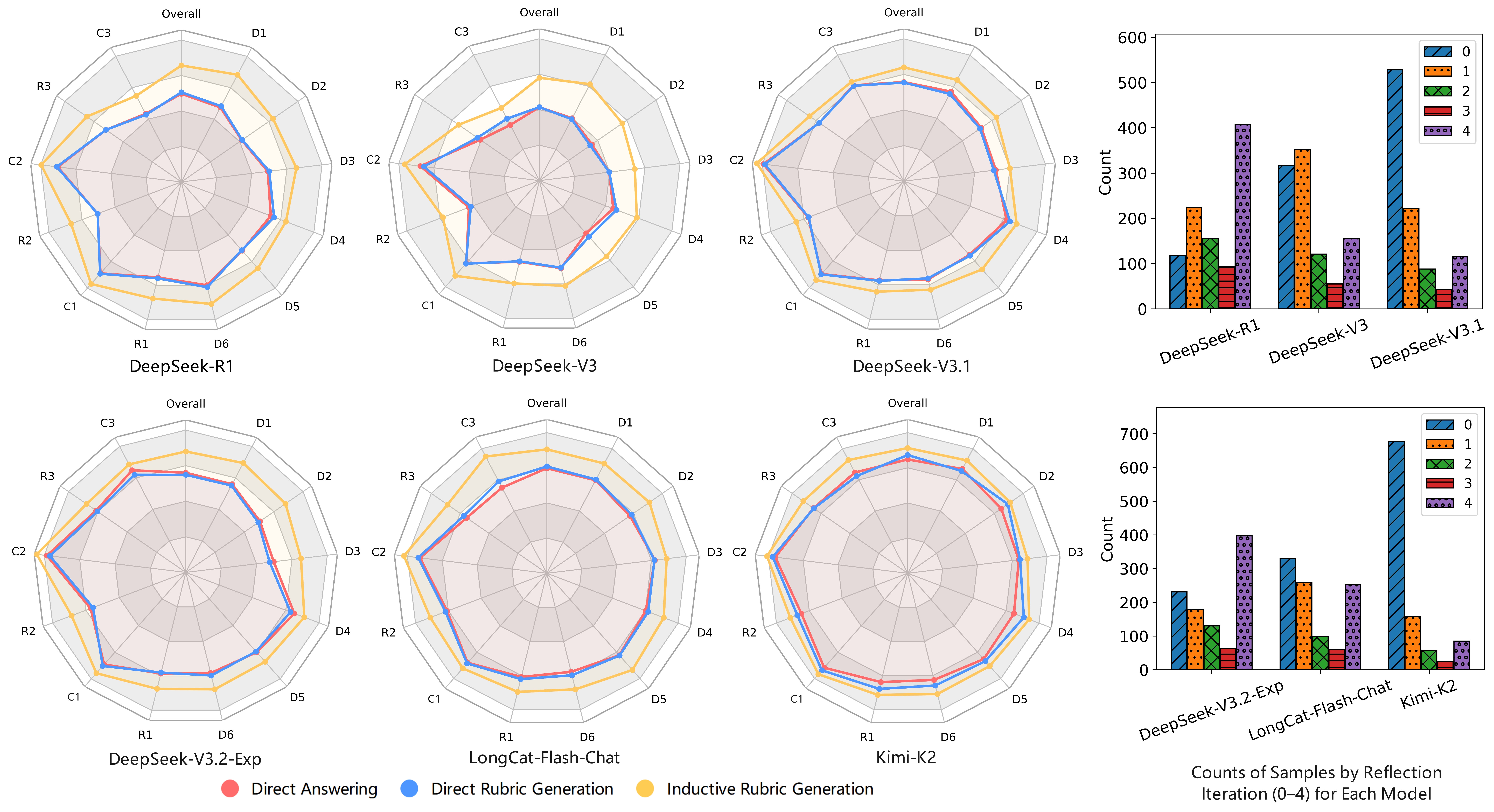}
    \caption{Rubric-guided revision with fixed writer parameters. Red: direct answering; blue: holistic LLM feedback; yellow: RubricGen-4B feedback. Radar plots report WritingBench scores; bars count samples requiring zero to four revisions. DeepSeek-R1: DeepSeek-R1-0528. D1 to D6: Academic and Engineering, Finance and Business, Politics and Law, Literature and Art, Education, and Advertising and Marketing. R1 to R3: Style, Format, and Length; C1 to C3: corresponding category scores. Overall: aggregate score.}
    \label{fig:leidatu}
\end{figure}

Figure~\ref{fig:leidatu} compares direct answering, holistic LLM feedback, and feedback based on rubrics generated by RubricGen-4B across six target writers. The radar plots show that RubricGen-4B guidance generally improves scores across the reported domains and requirements. The resulting score gains support the downstream utility of RubricGen-4B criteria as revision feedback. Together with its supervised training on the mined corpus, this provides evidence that the specifications can support other writers with fixed parameters.

Mean revision counts in Figure~\ref{fig:leidatu}, ordered from left to right across the top row and then the bottom row of radar plots, are 2.45, 1.38, 0.99, 2.22, 1.65, and 0.68. The observed revision counts vary considerably across writers. Means below one reflect a mixture that includes initial drafts accepted without revision; individual requests can still require multiple revisions. These distributions characterize the additional revision rounds used by each writer under the reported stopping rule.

\subsection{Comparison of Rubric Sources}
\label{app:sources}
Table~\ref{criteria_source1} compares rubric sources for DeepSeek-V3, DeepSeek-V3.1, and DeepSeek-V3.2-Exp. Within each target block, the writing model and the evaluation setting are held fixed while the source of the feedback criteria changes. This comparison tests whether automatic construction preserves useful task information relative to human-curated guidance.

Kimi-K2 rubrics exceed the human-curated reference on all three targets, with scores of 0.809, 0.855, and 0.849. LongCat-Flash-Chat rubrics also exceed the reference on each target, scoring 0.809, 0.845, and 0.843. The advantage appears with multiple constructors, supporting the portability of the rubric representation across model sources. In these settings, automatically mined criteria provide competitive guidance despite being constructed by a model different from the target writer.

Downstream rubric utility varies across targets. DeepSeek-R1 rubrics score 0.788 on DeepSeek-V3 and 0.836 on DeepSeek-V3.2-Exp, below the human references of 0.803 and 0.838, but reach 0.846 on DeepSeek-V3.1. These differences indicate that downstream utility depends on the combination of constructor and writer rather than on the constructor alone. Transferability should therefore be assessed across several targets instead of being inferred from a single model pair.

\begin{table}[t]
\caption{Writing performance under different rubric sources on WritingBench, with the target writer and evaluation setting fixed within each block. D1 to D6: Academic and Engineering, Finance and Business, Politics and Law, Literature and Art, Education, and Advertising and Marketing. R1 to R3: Style, Format, and Length; C1 to C3: corresponding category scores.}
\label{criteria_source1}
\centering
\resizebox{\textwidth}{!}{%
\begin{tabular}{llccccccccccccc}
\toprule
\multirow{2}{*}{\textbf{Target writer}} &\multirow{2}{*}{\textbf{Rubric source}} & \multirow{2}{*}{\textbf{Overall}} & \multicolumn{6}{c}{\textbf{Domains}} & \multicolumn{6}{c}{\textbf{Requirements}}    \\
\cmidrule(lr){4-9}
\cmidrule(lr){10-15}
 & & & \textbf{D1} & \textbf{D2} & \textbf{D3} & \textbf{D4} &\textbf{D5} & \textbf{D6} & \textbf{R1} & \textbf{C1} &\textbf{R2} & \textbf{C2} & \textbf{R3} & \textbf{C3} \\
\midrule
\multirow{7}{*}{\textbf{DeepSeek-V3}} & \cellcolor{lighty}Human-curated & \cellcolor{lighty}0.803 & \cellcolor{lighty}0.815 & \cellcolor{lighty}0.799 & \cellcolor{lighty}0.790 & \cellcolor{lighty}0.806 & \cellcolor{lighty}0.800 & \cellcolor{lighty}0.814 & \cellcolor{lighty}0.809 & \cellcolor{lighty}0.851 & \cellcolor{lighty}0.804 & \cellcolor{lighty}0.868 & \cellcolor{lighty}0.794 & \cellcolor{lighty}0.762 \\
 & DeepSeek-R1 & 0.788 & 0.791 & 0.780 & 0.781 & 0.793 & 0.773 & 0.816 & 0.800 & 0.832 & 0.793 & 0.850 & 0.791 & 0.743 \\
 & DeepSeek-R1-0528 & 0.806 & 0.814 & 0.799 & 0.791 & 0.808 & 0.802 & 0.831 & 0.817 & 0.858 & 0.810 & 0.866 & 0.801 & 0.767 \\
 & DeepSeek-V3.2-Exp & 0.806 & 0.809 & 0.790 & 0.802 & 0.817 & 0.797 & 0.831 & 0.817 & 0.851 & 0.811 & 0.868 & 0.806 & 0.761 \\
 & LongCat-Flash-Chat & 0.809 & 0.809 & 0.794 & 0.806 & 0.815 & 0.809 & 0.830 & 0.817 & 0.854 & 0.813 & 0.875 & 0.804 & 0.760 \\
 & Kimi-K2 & 0.809 & 0.820 & 0.795 & 0.797 & 0.809 & 0.815 & 0.829 & 0.816 & 0.861 & 0.814 & 0.873 & 0.805 & 0.751 \\
 & \cellcolor{lightg}RubricGen-4B & \cellcolor{lightg}0.803 & \cellcolor{lightg}0.800 & \cellcolor{lightg}0.788 & \cellcolor{lightg}0.799 & \cellcolor{lightg}0.816 & \cellcolor{lightg}0.793 & \cellcolor{lightg}0.825 & \cellcolor{lightg}0.818 & \cellcolor{lightg}0.858 & \cellcolor{lightg}0.808 & \cellcolor{lightg}0.869 & \cellcolor{lightg}0.796 & \cellcolor{lightg}0.752 \\
\addlinespace[1pt]
\cdashline{1-15}
\addlinespace[1pt]
\multirow{7}{*}{\textbf{DeepSeek-V3.1}} & \cellcolor{lighty}Human-curated & \cellcolor{lighty}0.824 & \cellcolor{lighty}0.825 & \cellcolor{lighty}0.821 & \cellcolor{lighty}0.810 & \cellcolor{lighty}0.837 & \cellcolor{lighty}0.832 & \cellcolor{lighty}0.820 & \cellcolor{lighty}0.824 & \cellcolor{lighty}0.860 & \cellcolor{lighty}0.826 & \cellcolor{lighty}0.891 & \cellcolor{lighty}0.825 &\cellcolor{lighty} 0.821 \\
 & DeepSeek-R1 & 0.846 & 0.846 & 0.854 & 0.839 & 0.847 & 0.845 & 0.848 & 0.846 & 0.869 & 0.851 & 0.893 & 0.841 & 0.824 \\
 & DeepSeek-R1-0528 & 0.834 & 0.842 & 0.829 & 0.817 & 0.843 & 0.839 & 0.840 & 0.836 & 0.867 & 0.836 & 0.890 & 0.837 & 0.838 \\
 & DeepSeek-V3.2-Exp & 0.842 & 0.850 & 0.843 & 0.826 & 0.849 & 0.848 & 0.841 & 0.842 & 0.868 & 0.844 & 0.890 & 0.838 & 0.815 \\
 & LongCat-Flash-Chat & 0.845 & 0.849 & 0.844 & 0.833 & 0.848 & 0.850 & 0.855 & 0.846 & 0.873 & 0.849 & 0.895 & 0.844 & 0.839 \\
 & Kimi-K2 & 0.855 & 0.860 & 0.854 & 0.846 & 0.858 & 0.853 & 0.858 & 0.856 & 0.884 & 0.856 & 0.902 & 0.850 & 0.827 \\
 & \cellcolor{lightg}RubricGen-4B & \cellcolor{lightg}0.846 & \cellcolor{lightg}0.854 & \cellcolor{lightg}0.842 & \cellcolor{lightg}0.834 & \cellcolor{lightg}0.850 & \cellcolor{lightg}0.850 & \cellcolor{lightg}0.853 & \cellcolor{lightg}0.847 & \cellcolor{lightg}0.876 & \cellcolor{lightg}0.850 & \cellcolor{lightg}0.891 & \cellcolor{lightg}0.850 & \cellcolor{lightg}0.833 \\
\addlinespace[1pt]
\cdashline{1-15}
\addlinespace[1pt]
\multirow{7}{*}{\textbf{DeepSeek-V3.2-Exp}}&\cellcolor{lighty}Human-curated & \cellcolor{lighty}0.838 & \cellcolor{lighty}0.843 & \cellcolor{lighty}0.838 & \cellcolor{lighty}0.828 & \cellcolor{lighty}0.849 & \cellcolor{lighty}0.835 & \cellcolor{lighty}0.837 & \cellcolor{lighty}0.836 & \cellcolor{lighty}0.865 & \cellcolor{lighty}0.840 & \cellcolor{lighty}0.895 & \cellcolor{lighty}0.837 & \cellcolor{lighty}0.840 \\
 & DeepSeek-R1 & 0.836 & 0.844 & 0.834 & 0.831 & 0.841 & 0.826 & 0.840 & 0.840 & 0.865 & 0.842 & 0.886 & 0.837 & 0.806 \\
 & DeepSeek-R1-0528 & 0.841 & 0.847 & 0.838 & 0.826 & 0.844 & 0.847 & 0.850 & 0.844 & 0.873 & 0.845 & 0.895 & 0.841 & 0.835 \\
 & DeepSeek-V3.2-Exp & 0.840 & 0.847 & 0.839 & 0.822 & 0.849 & 0.837 & 0.848 & 0.842 & 0.870 & 0.845 & 0.892 & 0.840 & 0.811 \\
 & LongCat-Flash-Chat & 0.843 & 0.848 & 0.840 & 0.833 & 0.841 & 0.846 & 0.855 & 0.844 & 0.870 & 0.846 & 0.894 & 0.844 & 0.827 \\
 & Kimi-K2 & 0.849 & 0.856 & 0.850 & 0.837 & 0.849 & 0.852 & 0.854 & 0.852 & 0.879 & 0.851 & 0.897 & 0.850 & 0.832 \\
 & \cellcolor{lightg}RubricGen-4B & \cellcolor{lightg}0.864 & \cellcolor{lightg}0.853 & \cellcolor{lightg}0.857 & \cellcolor{lightg}0.859 & \cellcolor{lightg}0.878 & \cellcolor{lightg}0.872 & \cellcolor{lightg}0.873 & \cellcolor{lightg}0.867 & \cellcolor{lightg}0.889 & \cellcolor{lightg}0.867 & \cellcolor{lightg}0.902 & \cellcolor{lightg}0.860 & \cellcolor{lightg}0.820 \\
\bottomrule
\end{tabular}%
}
\end{table}

\subsection{RubricGen-4B: Training and Downstream Utility}
\label{app:generator}
RubricGen-4B is trained by SFT on all 19K prompt and rubric pairs produced by Qwen3-32B through IRG. The target sequence contains the dimensions, scope descriptions, and scoring descriptions produced by the mining procedure. This direct mapping transfers the outputs of IRG to a compact model that produces a complete rubric for each new request.

Responses revised with RubricGen-4B criteria match the human-curated reference on DeepSeek-V3 at 0.803 and exceed it on DeepSeek-V3.1 by 0.022, reaching 0.846. With DeepSeek-V3.2-Exp, they score 0.864, exceeding the scores obtained with human-curated and Kimi-K2 rubrics by 0.026 and 0.015, respectively. The advantage is not uniform across evaluation metrics: on the last target, responses guided by RubricGen-4B have a lower length category score than those guided by human-curated rubrics. The overall scores support the effectiveness of the mined corpus as supervision for compact rubric generation.

The broader comparison in Figure~\ref{fig:rubric-transfer} shows higher scores with RubricGen-4B guidance than with direct answering for every displayed target. Together, these results support using the mined corpus to train a generator whose criteria guide responses across writers.

\section{Evaluator Robustness and Human Assessment}
\label{app:evaluation}
The evaluator study in Section~\ref{sec:evaluators} uses the WritingBench LLM evaluation setting with DeepSeek-V3.2-Exp, Qwen3-235B-A22B, GLM-5.1, and Kimi-K2. Human assessment uses four experts, each of whom independently evaluates the anonymized responses. Model identities are hidden, and all experts follow the same benchmark criteria, score scale, and aggregation rules used for LLM assessment. These evaluations assess the writing outputs separately from the main critic-based protocol, so the same responses are judged under several distinct assessment configurations. Figure~\ref{fig:evaluator-transfer} reports the overall comparison, and Table~\ref{criteria_llm1} provides domain and requirement scores for three LLM evaluators. Because the assessment protocols differ, their absolute scores are not directly comparable to the critic-based scores in Table~\ref{main-performance1}; robustness is therefore assessed through comparisons within each evaluation setting rather than across settings.

For each expert, criterion-level judgments are aggregated with the benchmark scoring procedure. The reported human result is the arithmetic mean of the four independently obtained scores,
\begin{equation}
\label{eq:human-score}
S_{\mathrm{human}}=\frac{1}{4}\sum_{j=1}^{4}S_{\mathrm{benchmark}}^{(j)}.
\end{equation}
The same averaging applies to the corresponding domain and requirement summaries. This protocol gives each expert equal weight while preserving the evaluation criteria and aggregation used for the model-based comparison.

Writer-R1-8B ranks first overall under all five sources. Both variants also improve over their backbones in every domain reported for the three evaluators in Table~\ref{criteria_llm1}. Human assessment places Writer-R1-8B at 0.741 and Writer-R1-4B at 0.708, ahead of LongWriter-Zero-32B at 0.684. Consistent improvements over the backbones across domains and evaluation sources strengthen the evidence that the training gains extend across assessment settings.

Some pairwise rankings remain evaluator-dependent. GLM-5.1 scores Writer-R1-4B at 0.598, below LongWriter-Zero-32B at 0.606, whereas the other four sources rank Writer-R1-4B higher. All sources rank both Writer-R1 variants above their respective backbones. These pairwise differences qualify the ranking of the 4B model while preserving the consistent improvements over both backbones.

\begin{table}[t]
\caption{Domain and requirement performance on WritingBench under different LLM evaluators. D1 to D6: Academic and Engineering, Finance and Business, Politics and Law, Literature and Art, Education, and Advertising and Marketing. R1 to R3: Style, Format, and Length; C1 to C3: corresponding category scores.}
\label{criteria_llm1}
\centering
\resizebox{\textwidth}{!}{%
\begin{tabular}{llccccccccccccc}
\toprule
\multirow{2}{*}{\textbf{Evaluator}} &\multirow{2}{*}{\textbf{Writing model}} & \multirow{2}{*}{\textbf{Overall}} & \multicolumn{6}{c}{\textbf{Domains}} & \multicolumn{6}{c}{\textbf{Requirements}}    \\
\cmidrule(lr){4-9}
\cmidrule(lr){10-15}
 & & & \textbf{D1} & \textbf{D2} & \textbf{D3} & \textbf{D4} &\textbf{D5} & \textbf{D6} & \textbf{R1} & \textbf{C1} &\textbf{R2} & \textbf{C2} & \textbf{R3} & \textbf{C3} \\
\midrule
\multirow{7}{*}{\textbf{DeepSeek-V3.2-Exp}} & Qwen3-4B & 0.523 & 0.541 & 0.517 & 0.538 & 0.461 & 0.561 & 0.544 & 0.535 & 0.589 & 0.532 & 0.615 & 0.498 & 0.493 \\
 & Qwen3-8B & 0.573 & 0.583 & 0.568 & 0.570 & 0.535 & 0.605 & 0.598 & 0.583 & 0.634 & 0.579 & 0.664 & 0.555 & 0.558 \\
 & WritingBench-7B & 0.614 & 0.641 & 0.585 & 0.614 & 0.573 & 0.648 & 0.655 & 0.630 & 0.675 & 0.611 & 0.669 & 0.599 & 0.579 \\
 & SuperWriter-7B & 0.624 & 0.628 & 0.611 & 0.626 & 0.594 & 0.665 & 0.644 & 0.637 & 0.674 & 0.622 & 0.675 & 0.608 & 0.559 \\
 & LongWriter-Zero-32B & 0.703 & 0.710 & 0.693 & 0.696 & 0.682 & 0.740 & 0.720 & 0.704 & 0.739 & 0.702 & 0.744 & 0.697 & 0.700 \\
 & \cellcolor{lightg}Writer-R1-4B & \cellcolor{lightg}0.721 & \cellcolor{lightg}0.755 & \cellcolor{lightg}0.736 & \cellcolor{lightg}0.718 & \cellcolor{lightg}0.651 & \cellcolor{lightg}0.768 & \cellcolor{lightg}0.715 & \cellcolor{lightg}0.714 & \cellcolor{lightg}0.748 & \cellcolor{lightg}0.724 & \cellcolor{lightg}0.778 & \cellcolor{lightg}0.698 & \cellcolor{lightg}0.695 \\
 & \cellcolor{lightg}Writer-R1-8B & \cellcolor{lightg}0.747 & \cellcolor{lightg}0.786 & \cellcolor{lightg}0.750 & \cellcolor{lightg}0.743 & \cellcolor{lightg}0.708 & \cellcolor{lightg}0.771 & \cellcolor{lightg}0.734 & \cellcolor{lightg}0.742 & \cellcolor{lightg}0.770 & \cellcolor{lightg}0.754 & \cellcolor{lightg}0.796 & \cellcolor{lightg}0.737 & \cellcolor{lightg}0.717 \\
\addlinespace[1pt]
\cdashline{1-15}
\addlinespace[1pt]
\multirow{7}{*}{\textbf{GLM-5.1}} & Qwen3-4B & 0.479 & 0.516 & 0.491 & 0.492 & 0.388 & 0.508 & 0.496 & 0.484 & 0.527 & 0.494 & 0.569 & 0.449 & 0.399 \\
 & Qwen3-8B & 0.524 & 0.556 & 0.530 & 0.526 & 0.459 & 0.555 & 0.540 & 0.530 & 0.580 & 0.534 & 0.614 & 0.506 & 0.448 \\
 & WritingBench-7B & 0.538 & 0.576 & 0.522 & 0.551 & 0.467 & 0.568 & 0.568 & 0.543 & 0.589 & 0.542 & 0.593 & 0.515 & 0.420 \\
 & SuperWriter-7B & 0.543 & 0.580 & 0.537 & 0.561 & 0.470 & 0.582 & 0.546 & 0.548 & 0.580 & 0.542 & 0.590 & 0.515 & 0.415 \\
 & LongWriter-Zero-32B & 0.606 & 0.646 & 0.606 & 0.611 & 0.529 & 0.659 & 0.607 & 0.596 & 0.632 & 0.615 & 0.656 & 0.573 & 0.548 \\
 & \cellcolor{lightg}Writer-R1-4B & \cellcolor{lightg}0.598 & \cellcolor{lightg}0.649 & \cellcolor{lightg}0.630 & \cellcolor{lightg}0.598 & \cellcolor{lightg}0.480 & \cellcolor{lightg}0.663 & \cellcolor{lightg}0.587 & \cellcolor{lightg}0.584 & \cellcolor{lightg}0.619 & \cellcolor{lightg}0.610 & \cellcolor{lightg}0.652 & \cellcolor{lightg}0.552 & \cellcolor{lightg}0.472 \\
 & \cellcolor{lightg}Writer-R1-8B & \cellcolor{lightg}0.651 & \cellcolor{lightg}0.687 & \cellcolor{lightg}0.663 & \cellcolor{lightg}0.665 & \cellcolor{lightg}0.556 & \cellcolor{lightg}0.696 & \cellcolor{lightg}0.655 & \cellcolor{lightg}0.643 & \cellcolor{lightg}0.692 & \cellcolor{lightg}0.661 & \cellcolor{lightg}0.697 & \cellcolor{lightg}0.620 & \cellcolor{lightg}0.538 \\
\addlinespace[1pt]
\cdashline{1-15}
\addlinespace[1pt]
\multirow{7}{*}{\textbf{Kimi-K2}} & Qwen3-4B & 0.521 & 0.542 & 0.532 & 0.547 & 0.444 & 0.569 & 0.500 & 0.523 & 0.559 & 0.529 & 0.621 & 0.490 & 0.459 \\
 & Qwen3-8B & 0.568 & 0.571 & 0.577 & 0.589 & 0.516 & 0.618 & 0.550 & 0.573 & 0.612 & 0.574 & 0.657 & 0.554 & 0.530 \\
 & WritingBench-7B & 0.600 & 0.634 & 0.585 & 0.630 & 0.535 & 0.642 & 0.588 & 0.605 & 0.643 & 0.601 & 0.654 & 0.582 & 0.526 \\
 & SuperWriter-7B & 0.604 & 0.614 & 0.612 & 0.623 & 0.548 & 0.661 & 0.575 & 0.611 & 0.636 & 0.600 & 0.646 & 0.582 & 0.501 \\
 & LongWriter-Zero-32B & 0.665 & 0.675 & 0.670 & 0.687 & 0.610 & 0.722 & 0.640 & 0.660 & 0.696 & 0.668 & 0.701 & 0.652 & 0.667 \\
 & \cellcolor{lightg}Writer-R1-4B & \cellcolor{lightg}0.720 & \cellcolor{lightg}0.776 & \cellcolor{lightg}0.754 & \cellcolor{lightg}0.747 & \cellcolor{lightg}0.612 & \cellcolor{lightg}0.772 & \cellcolor{lightg}0.660 & \cellcolor{lightg}0.705 & \cellcolor{lightg}0.742 & \cellcolor{lightg}0.729 & \cellcolor{lightg}0.768 & \cellcolor{lightg}0.675 & \cellcolor{lightg}0.621 \\
 & \cellcolor{lightg}Writer-R1-8B & \cellcolor{lightg}0.745 & \cellcolor{lightg}0.770 & \cellcolor{lightg}0.744 & \cellcolor{lightg}0.760 & \cellcolor{lightg}0.657 & \cellcolor{lightg}0.764 & \cellcolor{lightg}0.714 & \cellcolor{lightg}0.744 & \cellcolor{lightg}0.790 & \cellcolor{lightg}0.755 & \cellcolor{lightg}0.792 & \cellcolor{lightg}0.747 & \cellcolor{lightg}0.665 \\
\bottomrule
\end{tabular}%
}
\end{table}

\section{Sensitivity to Revision and Decoding Parameters}
\label{app:sensitivity}
\subsection{Revision Budget and Acceptance Threshold}
\label{app:revision}
This appendix examines two sets of operating parameters: the revision budget and acceptance threshold that govern rubric-guided self-improvement with fixed model parameters, and the decoding parameters that govern generation by the trained Writer-R1-4B policy. The self-improvement procedure has two controls: the maximum number of generation-and-assessment rounds $R_{\max}$ and the acceptance threshold $\eta$. It stops when the normalized rubric score reaches $\eta$ or the budget is exhausted. Increasing $R_{\max}$ permits additional drafts, while increasing $\eta$ requires stronger rubric satisfaction before acceptance. The two controls interact because the revision budget limits how long a response can be improved toward the acceptance threshold.

\begin{figure}[t]
    \centering
    \includegraphics[width=0.62\textwidth]{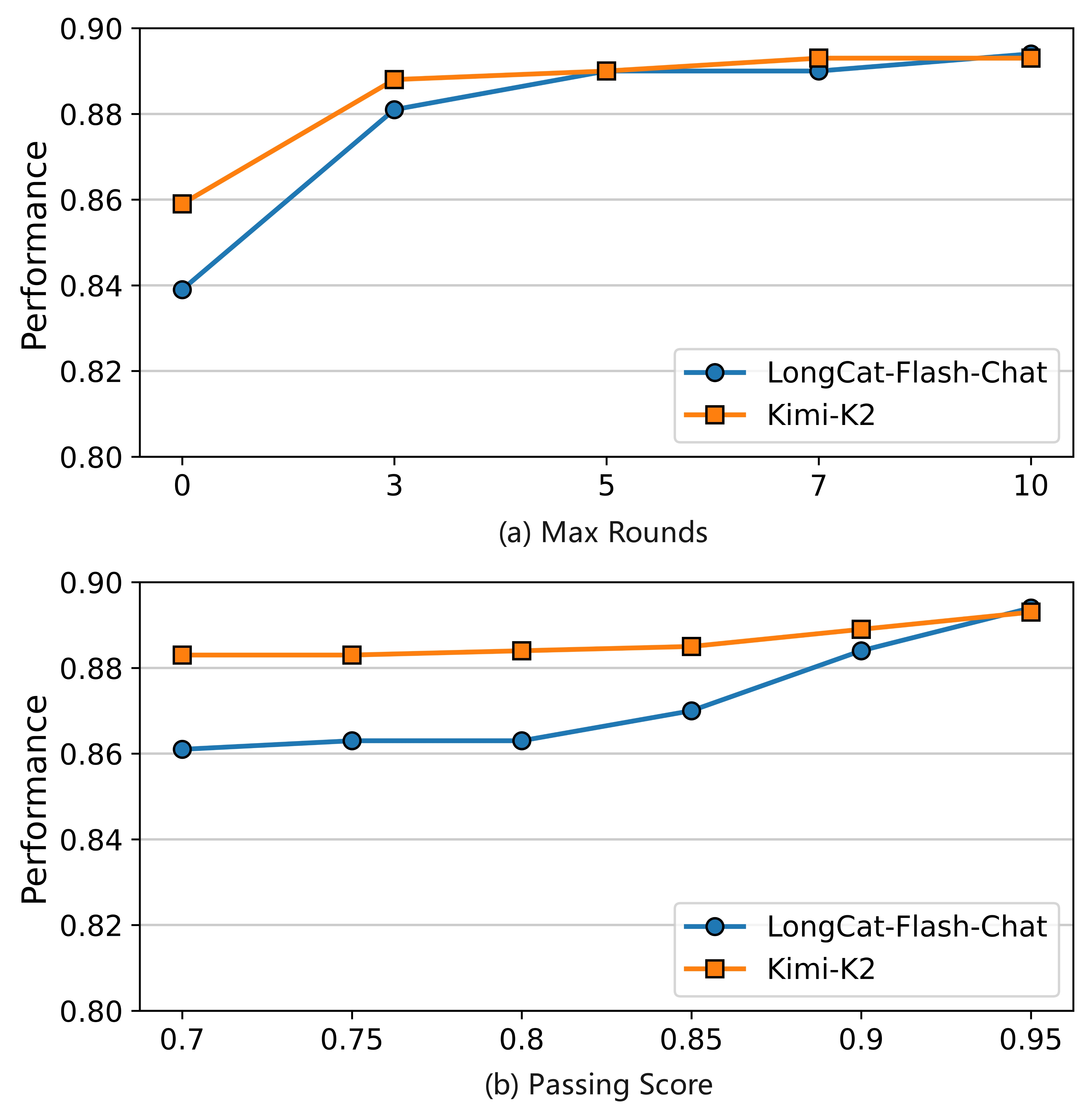}
    \caption{Sensitivity of writing performance to revision budget (a) and acceptance threshold (b) on WritingBench for LongCat-Flash-Chat and Kimi-K2. Max Rounds is $R_{\max}$, including the initial draft; zero indicates direct answering. Passing Score is the normalized threshold $\eta$.}
    \label{fig:revision-sensitivity}
\end{figure}

In Figure~\ref{fig:revision-sensitivity}(a), most of the improvement for both writers occurs within a budget of three rounds, with smaller gains beyond five. Figure~\ref{fig:revision-sensitivity}(b) shows a generally positive response to stricter acceptance thresholds, particularly for LongCat-Flash-Chat. Kimi-K2 is less sensitive at lower thresholds but also benefits from a high target score. These trends show that the criteria guide useful revisions within a bounded number of rounds, while later rounds yield smaller score improvements.

\subsection{Decoding Sensitivity}
\label{app:decoding}
The second study varies temperature and top-$p$ for Writer-R1-4B to test whether its benchmark performance depends on a narrow decoding configuration. Temperature adjusts the concentration of the sampling distribution, while top-$p$ restricts sampling to a nucleus of tokens with the specified cumulative probability mass.

\begin{figure}[t]
    \centering
    \includegraphics[width=0.62\textwidth]{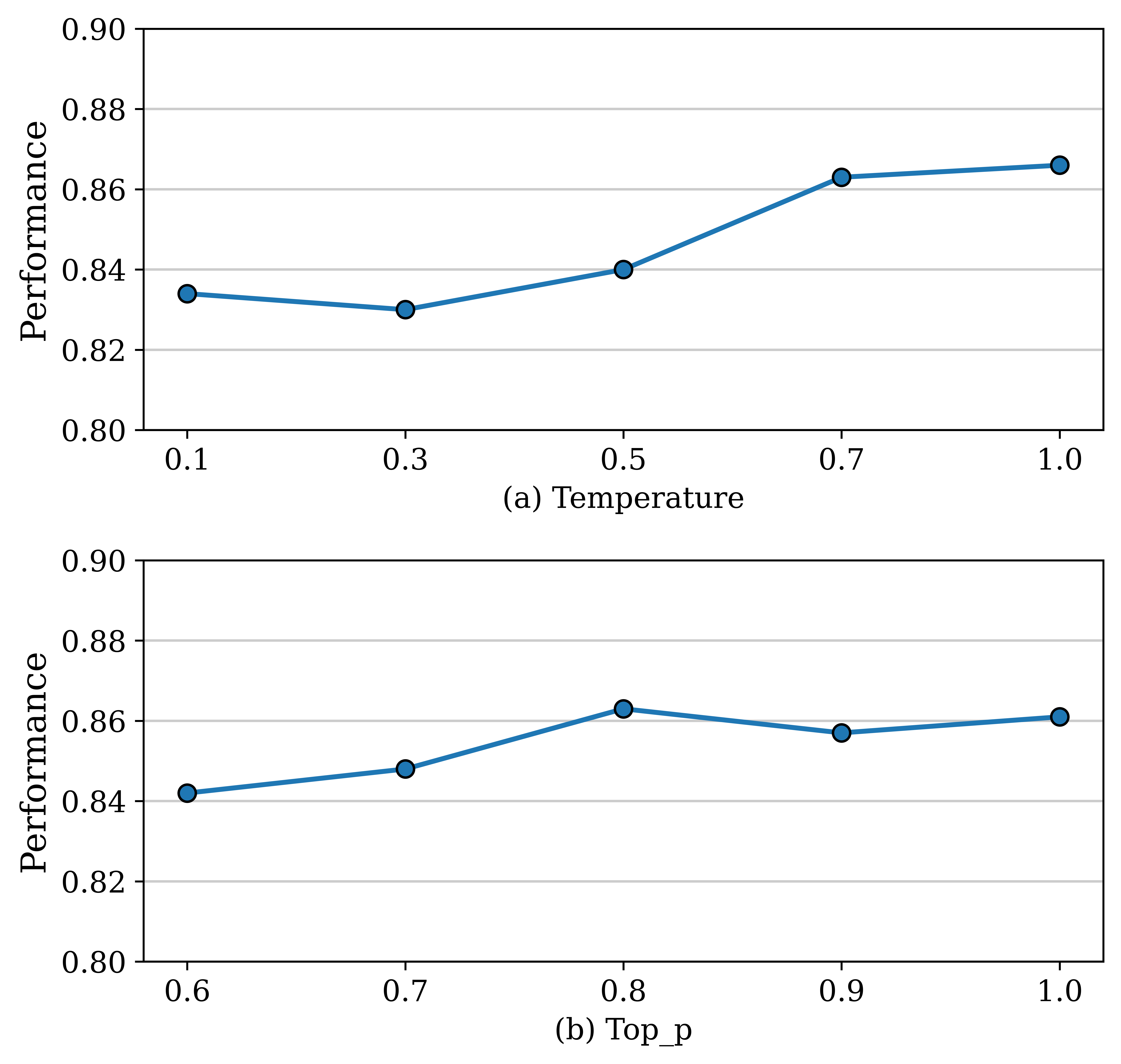}
    \caption{Sensitivity of Writer-R1-4B performance on WritingBench to sampling temperature (a) and top-$p$ (b).}
    \label{fig:decoding-sensitivity}
\end{figure}

Figure~\ref{fig:decoding-sensitivity}(a) shows stronger performance at the tested temperatures of 0.7 and 1.0 than at lower temperatures. The top-$p$ sweep in Figure~\ref{fig:decoding-sensitivity}(b) exhibits smaller score variations, with the best observed result at 0.8. These observations support the reported setting of temperature 0.7 and top-$p$ 0.8 within the tested range, with stronger sensitivity to temperature than to the nucleus threshold.

\section{AI Assistance Disclosure}

During the preparation of this manuscript, we used AI as an assistive tool for software development and academic writing. Specifically, it was used to assist with the implementation of parts of the proposed method and experimental pipeline, including drafting, revising, and debugging code. It was also used to assist with drafting and editing portions of the manuscript and to improve the clarity, readability, grammatical correctness, and overall quality of the English writing. All AI-generated or AI-assisted code was carefully reviewed, tested, and validated by the authors, and all AI-assisted text was reviewed and revised before inclusion in the manuscript.

Generative AI tools were not used to independently formulate the research questions or hypotheses, determine the research methodology or experimental design, interpret the experimental results, or derive the scientific conclusions of this work. All methodological decisions, experimental analyses, and scientific claims were made and verified by the authors. We take full responsibility for the correctness, originality, and final content of this work, including any text, code, or other material produced with the assistance of generative AI.

\section{Operational Prompt Templates}
\label{app:prompts}
This appendix collects English translations of the operational prompts and the benchmark scoring instruction. Placeholders denote fields inserted at execution time. The templates follow the order of data construction, rubric induction, inference, and assessment.

\subsection{Prompt Reconstruction and Enrichment Instructions}
\label{app:reconstruction-prompts}
Both reconstruction strategies use the same rewriting constraints. The second additionally invokes a length-recovery instruction when the reconstructed input is too short. The following panels translate the operational instructions into English.

\begin{promptbox}{Writing-prompt reconstruction}
You are a query rewriter. Given an original query, construct a new query that is semantically related but substantially different in content. Preserve the high-level task and difficulty, and match the total character length of the original query.
\begin{enumerate}
\setlength{\itemsep}{2pt}
\setlength{\parskip}{0pt}
    \item Reconstruct the request rather than lightly editing its sentences or paragraphs. Change its expression, organization, materials, details, and examples while preserving the underlying intent.
    \item If the query includes source material, a synopsis, an excerpt, data, or background text, generate new material. Keep it thematically related while substantially changing its setting, characters, places, and sequence of events.
    \item Preserve the type of requested output and comparable difficulty. Length, structural, and stylistic constraints may be adjusted while retaining that level of difficulty.
    \item Preserve the language. Keep Chinese requests in Chinese, English requests primarily in English, and approximately maintain the language proportions of mixed-language inputs.
    \item Produce a clear writing instruction that can be used directly as a new query. Output only the reconstructed request, without explanations or commentary.
\end{enumerate}
\textbf{Original query}\par
\texttt{\{query\}}
\par\medskip
Review the reconstruction rules before responding. Return only the final rewritten query, matching the total character length of the original.
\end{promptbox}

\begin{promptbox}{Length recovery and contextual enrichment}
You are a query expander. Extend the supplied query into a longer instruction for long-context model training.
\begin{enumerate}
\setlength{\itemsep}{2pt}
\setlength{\parskip}{0pt}
    \item Preserve its core intent, objective, and direction. Expand or clarify details without introducing unrelated tasks.
    \item Make the expanded query substantially longer than the supplied version.
    \item Keep the added content natural, reasonable, and internally consistent.
    \item Preserve the language and the approximate language proportions of mixed-language requests.
    \item Output only the expanded query, without explanations.
\end{enumerate}
\textbf{When source material is present.}
If the query includes a passage, synopsis, data, background, case, dialogue, transcript, minutes, log, table, or other supplied context, comprehensively rewrite and extend that material. Produce fuller and longer material, retaining the task intent while enriching details, settings, entities, events, data points, and contextual relationships. Integrate it naturally into the instruction.
\par\medskip
\textbf{When source material is absent.}
Expand the existing instruction moderately. Do not introduce assumptions that change its direction or invent a large body of unrelated material. Improve specificity and executability while preserving its purpose.
\par\medskip
\textbf{Query to expand}\par
\texttt{\{rewrite\_query\}}
\par\medskip
Review these requirements and return only the expanded query, which must be longer than the supplied version.
\end{promptbox}

\subsection{IRG Prompt Templates}
\label{app:irg-prompts}
IRG uses separate discovery and consolidation calls, followed by one scoring-specification call per retained dimension. The latter calls can run in parallel. Discovery and consolidation return tagged dimension names and rationales; the retained rationale is stored as the assessment scope under \texttt{criteria\_description}. The following panels translate the three operational prompts.

\begin{promptbox}{IRG stage 1: requirement discovery}
You are an evaluation-dimension generator for creative writing tasks. Given a user query, produce a comprehensive set of actionable and distinguishable evaluation dimensions. Each dimension must contain:
\par\smallskip
\texttt{name}: a concise, clear dimension name.\par
\texttt{reason}: what the dimension assesses, why it matters, and how it can be observed or checked.
\par\medskip
Apply the following principles:
\begin{enumerate}
\setlength{\itemsep}{2pt}
\setlength{\parskip}{0pt}
    \item Match the topic, genre, audience, tone, and scenario of the query. Do not supply only generic writing criteria.
    \item Make at least half of the dimensions specific to this query or closely related tasks. Assess its distinctive subject matter, intended perspective, essential information, and implicit requirements.
    \item The remaining dimensions may cover broader qualities, such as expression and organization, but must be linked to the task objectives.
    \item Every dimension must support human or automated assessment. Use observable checkpoints rather than vague impressions.
    \item Produce 15 to 20 dimensions. Avoid synonymous repetition and substantial overlap, and give each dimension a clearly distinguishable assessment target.
\end{enumerate}
Output only repeated two-line records, with a blank line between records:
\par\smallskip
\texttt{name: <dimension name>}\par
\texttt{reason: <scope, importance, and observable evidence>}
\par\medskip
Do not add explanations, headings, numbering, bullets, or surrounding text.
\par\medskip
\textbf{User query}\par
\texttt{\{query\}}
\end{promptbox}

\begin{promptbox}{IRG stage 2: dimension consolidation}
You are a selector and consolidator of evaluation dimensions. Given candidate dimensions with their rationales and a user query, form the core set of evaluation dimensions.
\par\medskip
\textbf{Retention and consolidation}
\begin{itemize}
\setlength{\itemsep}{2pt}
\setlength{\parskip}{0pt}
    \item Preserve the core meaning of a candidate that is semantically complete, clearly delimited, and uniquely relevant to the query. Its wording may be improved, but do not introduce a new assessment target.
    \item Consolidate candidates that share assessment objects, standards, or areas of emphasis when they can form a stronger, more stable dimension.
    \item Preserve the essential information of consolidated candidates through a unified assessment rationale. Avoid concatenating concepts or listing the original dimensions.
    \item Give each consolidated dimension a concise, informative name.
\end{itemize}
\textbf{Final-set requirements}
\begin{itemize}
\setlength{\itemsep}{2pt}
\setlength{\parskip}{0pt}
    \item Match the query topic, genre, audience, tone, and scenario. At least half of the dimensions must be query-specific.
    \item Broader dimensions must remain connected to the task objectives. All dimensions must be assessable through observable properties.
    \item Return 5 to 8 complementary dimensions covering the core quality requirements. Avoid synonymous repetition and substantial overlap.
\end{itemize}
Output only repeated two-line records with \texttt{name} followed by \texttt{reason}, separated by blank lines. Do not include headings, numbering, bullets, or surrounding commentary.
\par\medskip
\textbf{Candidate dimensions}\par
\texttt{\{metrics\_block\}}
\par\smallskip
\textbf{User query}\par
\texttt{\{query\}}
\end{promptbox}

\begin{promptbox}{IRG stage 3: scoring specification}
You are a scoring-rubric generator. For the supplied evaluation dimension and assessment scope, construct comprehensive, actionable, and discriminative score descriptions for consistent evaluation of responses to the user query.
\par\medskip
\textbf{Dimension name}\par
\texttt{\{name\}}
\par\smallskip
\textbf{Assessment scope}\par
\texttt{\{criteria\_description\}}
\par\medskip
Generate 5 to 8 intervals spanning clear failure through full or exceptional satisfaction. The maximum score need not be 10. Each interval must contain two consecutive integer scores, and the intervals must be consecutive, non-overlapping, and strictly increasing.
\par\medskip
For every interval, keep the description consistent with the assessment scope. Use observable, verifiable properties and concrete evidence where possible. Progressively increase coverage, depth, and satisfaction of constraints. Distinguish adjacent intervals clearly enough for consistent judgments.
\par\medskip
Output only interval labels and their descriptions, one interval per line:
\par\smallskip
\texttt{1-2: <performance description>}\par
\texttt{3-4: <performance description>}\par
\texttt{5-6: <performance description>}\par
Continue in the same format for the remaining intervals. Do not include headings or additional explanations.
\par\medskip
\textbf{User query, for task context}\par
\texttt{\{query\}}
\end{promptbox}

\subsection{RubricGen-4B Inference Prompt}
\label{app:rubricgen-prompt}
The following template translates the Chinese inference instruction in the RubricGen-4B generation script. The trained generator receives this instruction with the user query and returns a complete rubric in one call.

\begin{promptbox}{RubricGen-4B rubric generation}
You are a generator of evaluation rubrics for model responses. Given a user query, construct a professional, rigorous, actionable, and discriminative rubric for evaluating responses to that query. Analyze the task type, domain, implicit goals, constraints, expected output, characteristics of strong responses, and common deficiencies.
\par\medskip
\textbf{Dimension design}
\begin{enumerate}
\setlength{\itemsep}{2pt}
\setlength{\parskip}{0pt}
    \item Match the specific topic, genre, audience, tone, and scenario. Avoid criteria that merely reproduce a generic writing template.
    \item Make at least half of the dimensions specific to the query or closely related tasks. These dimensions should assess distinctive subject matter, perspectives, essential information, and implicit requirements.
    \item The remaining dimensions may address broader properties such as expression or organization, but must remain connected to the task objectives.
    \item Cover the essential properties of a high-quality response. Keep dimensions complementary and minimize repeated assessment of the same issue.
    \item Use finer, domain-specific dimensions for complex, professional, or research-oriented requests. For simpler requests, retain a moderate number while covering the core requirements.
    \item Usually generate 5 to 8 dimensions. Do not add redundant, vague, or poorly delimited criteria merely to reach a count.
\end{enumerate}
\textbf{Scope descriptions}
\par
For each dimension, ensure that \texttt{criteria\_description} corresponds precisely to \texttt{name}. State which requirements the dimension assesses and connect them to the particular query.
\par\medskip
\textbf{Scoring and output}
\par\smallskip
For each dimension, generate 5 to 8 score intervals spanning clear failure through full or exceptional satisfaction. The maximum need not be 10. Each interval must contain two consecutive integer scores, such as 1 to 2 and 3 to 4. Intervals must be consecutive, non-overlapping, and strictly increasing.
\par\medskip
Each performance description must:
\begin{itemize}
\setlength{\itemsep}{2pt}
\setlength{\parskip}{0pt}
    \item Remain consistent with the dimension scope.
    \item Use observable and verifiable properties, with concrete assessment points or characteristic evidence where possible.
    \item Increase progressively in coverage, depth, and satisfaction of constraints.
    \item Distinguish adjacent levels clearly enough for consistent assessment.
\end{itemize}
\textbf{Output schema}
\par
Return only a JSON list of dimension objects. Each object contains:
\par\smallskip
\begin{tabular}{@{}lp{0.62\linewidth}@{}}
\texttt{name} & Dimension name. \\
\texttt{criteria\_description} & Task-specific assessment scope. \\
\texttt{1-2}, \texttt{3-4}, $\ldots$ & Descriptions for successive score intervals. \\
\end{tabular}
\par\medskip
Do not include explanations or additional symbols outside the list.
\par\medskip
\textbf{User query}\par
\texttt{\{query\}}
\par\medskip
Analyze the query carefully and output the rubric.
\end{promptbox}

\subsection{Rubric-Based Reward Evaluation Prompt}
\label{app:reward-prompt}
The following panel translates the Chinese scoring instruction used for rubric-based response assessment. Field labels are rendered in English; reward parsing and aggregation are described in Appendix~\ref{app:optimization}.

\begin{promptbox}{Rubric-based response assessment}
You are a rigorous academic reviewer. Evaluate and comment on the model response item by item using the supplied rubric.
\par\medskip
Each rubric item contains a dimension name, an assessment scope, and descriptions for its score intervals. Assess every item in the order provided. For each item, return exactly three lines:
\par\smallskip
\texttt{name: <exact dimension name>}\par
\texttt{score: <obtained score>/<maximum score>}\par
\texttt{comments and suggestions: <assessment and improvements>}
\par\medskip
Continue until all items have been assessed. Use integer scores and numerical values in the score fields. Comments must identify both strengths and directions for improvement. Output only the repeated three-line structure; blank lines may separate items.
\par\medskip
\textbf{Model response}\par
\texttt{\{response\}}
\par\medskip
\textbf{Rubric}\par
\texttt{\{checklist\_str\}}
\end{promptbox}

\subsection{Rubric-Guided Revision Prompt}
\label{app:revision-prompt}
Assessment uses the dimension-wise scoring instruction above. The following panel translates the Chinese revision instruction. The previous review supplies criterion-specific feedback, while the original request remains the basis for the rewrite.

\begin{promptbox}{Rubric-guided draft revision}
You are a professional Chinese-language writing assistant. Read the original writing request and review comments carefully. Comprehensively rewrite and refine the previous response to produce an improved, complete text.
\begin{itemize}
\setlength{\itemsep}{2pt}
\setlength{\parskip}{0pt}
    \item Reconsider the overall structure and logical organization in light of the review, rather than limiting changes to local wording.
    \item Retain sound, high-quality points where appropriate, reorganizing and expressing them more effectively.
    \item Give particular attention to the weaknesses identified in the review.
    \item Return only the complete revised text, without additional explanations or commentary.
\end{itemize}
\textbf{Original writing request}\par
\texttt{\{query\}}
\par\smallskip
\textbf{Previous response}\par
\texttt{\{previous\_response\}}
\par\smallskip
\textbf{Review comments}\par
\texttt{\{previous\_review\}}
\par\medskip
Output only the final rewritten text.
\end{promptbox}

\subsection{Benchmark Scoring Instruction}
\label{app:benchmark-prompt}
The WritingBench evaluator receives the benchmark query, its evaluation criterion, and a candidate response. It returns an integer score from 1 to 10 with a justification. The scoring instruction below is typeset from the benchmark evaluation template; its criteria are supplied by the benchmark, separately from the rubric used to condition Writer-R1 generation.

\begin{promptbox}{Benchmark response evaluation}
You are an expert evaluator of responses to user queries. Evaluate the response against the supplied query and criterion using the following score levels:
\begin{itemize}
\setlength{\itemsep}{2pt}
\setlength{\parskip}{0pt}
    \item \textbf{1 to 2:} Critical deficiencies prevent an adequate response.
    \item \textbf{3 to 4:} Noticeable shortcomings impair effectiveness and require improvement.
    \item \textbf{5 to 6:} The response adequately meets essential requirements without exemplary execution.
    \item \textbf{7 to 8:} Strong execution with minor refinements needed.
    \item \textbf{9 to 10:} Exceptional execution that fully addresses the requirements without deficiencies.
\end{itemize}
Justify the score using specific strengths or weaknesses and relevant passages from the response. Evaluate strictly. Do not equate length or formatting with substantive quality. Identify fabricated content and distinguish a complete answer from an introduction or overview that fails to fulfill the request.
\par\medskip
\textbf{Criterion}\par
\texttt{\{criteria\}}
\par\smallskip
\textbf{Query}\par
\texttt{\{query\}}
\par\smallskip
\textbf{Response}\par
\texttt{\{response\}}
\par\medskip
Return only a JSON object containing \texttt{score}, an integer from 1 to 10, and \texttt{reason}, a specific explanation grounded in the response.
\end{promptbox}

\end{document}